\documentclass[10pt,twocolumn,letterpaper]{article}

\usepackage{cvpr}              

\usepackage{graphicx}
\usepackage{amssymb}
\usepackage{amsfonts}
\usepackage{amsmath}
\usepackage{mathrsfs}
\usepackage{makecell}
\usepackage{multirow}
\usepackage{multicol}
\usepackage{booktabs}
\usepackage{cuted}
\usepackage{capt-of}
\usepackage{placeins}
\usepackage{stackengine} 
\usepackage{indentfirst} 
\usepackage{pifont}
\usepackage{bm}
\usepackage{array}
\usepackage[accsupp]{axessibility}
\usepackage{color}
\definecolor{darkgreen}{rgb}{0,0.5,0.7}
\definecolor{or}{rgb}{1,0.5,0.25}

\newcommand\oast{\stackMath\mathbin{\stackinset{c}{0ex}{c}{0ex}{\ast}{\bigcirc}}}
\newcommand{\cmark}{\ding{51}}%
\newcommand{\xmark}{\ding{55}}%

\usepackage{tikz}

\definecolor{cvprblue}{rgb}{0.21,0.49,0.74}
\usepackage[pagebackref,breaklinks,colorlinks,citecolor=cvprblue]{hyperref}

\def\paperID{9217} 
\def\confName{CVPR}
\def\confYear{2024}

\title{FMA-Net: Flow-Guided Dynamic Filtering and Iterative Feature Refinement with Multi-Attention for Joint Video Super-Resolution and Deblurring}

\author{Geunhyuk Youk \qquad\qquad Munchurl Kim\thanks{Corresponding author.}\\
[0.5em]
Korea Advanced Institute of Science and Technology\\
{\tt\small \{rmsgurkjg, mkimee\}@kaist.ac.kr}
}

\author{Geunhyuk Youk\\
[0.3em]
KAIST\\
{\tt\small rmsgurkjg@kaist.ac.kr}
\and
Jihyong Oh \footnotemark[2]\\
[0.3em]
Chung-Ang University\\
{\tt\small jihyongoh@cau.ac.kr}
\and
Munchurl Kim \footnotemark[2]\\
[0.3em]
KAIST\\
{\tt\small mkimee@kaist.ac.kr}
}

\begin{document}

\twocolumn[{%
\renewcommand\twocolumn[1][]{#1}%
\maketitle
\begin{center}\centering
    \setlength{\tabcolsep}{0.06cm}
    \setlength{\columnwidth}{5.7cm}
    \hspace*{-\tabcolsep}\begin{tabular}{cc}
            \includegraphics[width=10cm]{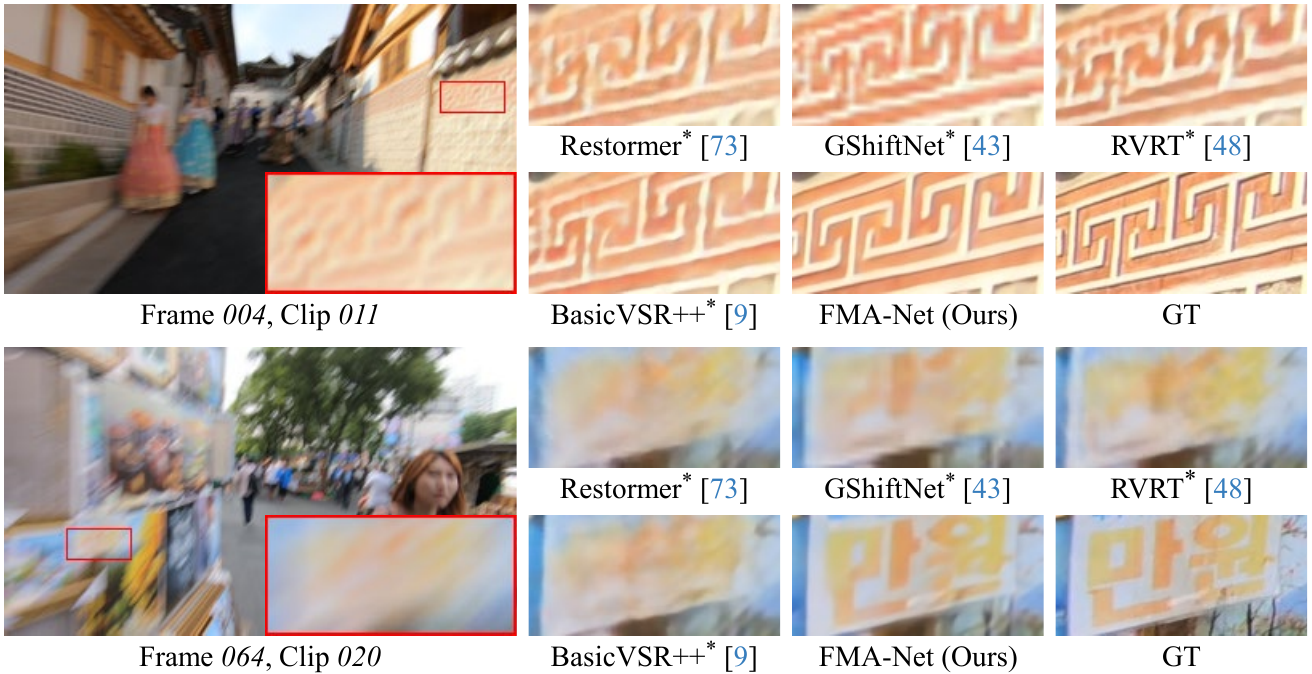}
        & \hspace{3mm}
            \includegraphics[width=6.25cm]{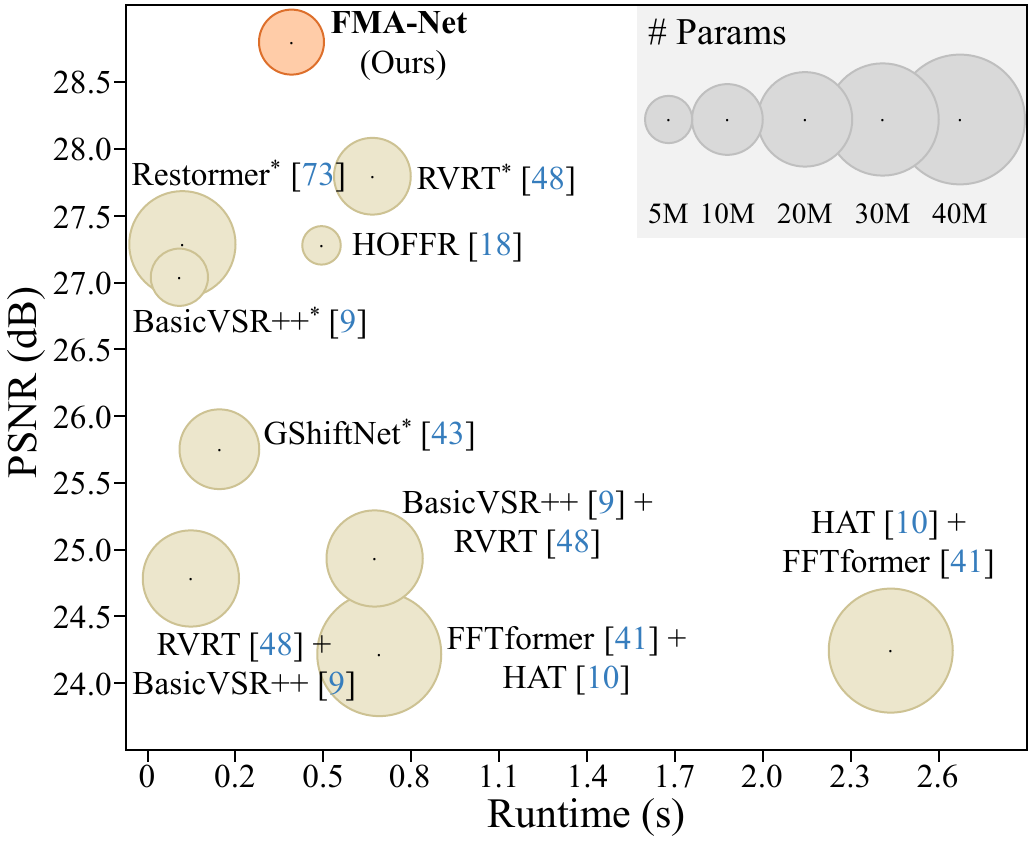}
        \\
            \footnotesize (a) Visual comparison results of different methods on REDS4 \cite{nah2019ntire} test set
        &
            \footnotesize (b) Performance Gain
        \\
        \\
    \end{tabular}\vspace{-0.4cm}
	\captionof{figure}{Our FMA-Net outperforms state-of-the-art methods in both quantitative and qualitative results for $\times 4$ VSRDB.}\vspace{-0.0cm}
	\label{fig:teaser}
\end{center}
}]

\begin{abstract}
\vspace{-4mm}
We present a joint learning scheme of video super-resolution and deblurring, called VSRDB, to restore clean high-resolution (HR) videos from blurry low-resolution (LR) ones. This joint restoration problem has drawn much less attention compared to single restoration problems. In this paper, we propose a novel flow-guided dynamic filtering (FGDF) and iterative feature refinement with multi-attention (FRMA), which constitutes our VSRDB framework, denoted as FMA-Net. Specifically, our proposed FGDF enables precise estimation of both spatio-temporally-variant degradation and restoration kernels that are aware of motion trajectories through sophisticated motion representation learning. Compared to conventional dynamic filtering, the FGDF enables the FMA-Net to effectively handle large motions into the VSRDB. Additionally, the stacked FRMA blocks trained with our novel temporal anchor (TA) loss, which temporally anchors and sharpens features, refine features in a coarse-to-fine manner through iterative updates. Extensive experiments demonstrate the superiority of the proposed FMA-Net over state-of-the-art methods in terms of both quantitative and qualitative quality. Codes and pre-trained models are available at: \url{https://kaist-viclab.github.io/fmanet-site}.
\end{abstract}
\enlargethispage{\baselineskip}
{
  \renewcommand{\thefootnote}%
    {\fnsymbol{footnote}}
  \footnotetext[2]{Co-corresponding authors.}
}
\vspace{-5mm}
\section{Introduction}
Video super-resolution (VSR) aims to restore a high-resolution (HR) video from a given low-resolution (LR) counterpart. VSR can be beneficial for diverse real-world applications of high-quality video, such as surveillance \cite{zhang2010super, aakerberg2022real}, video streaming \cite{dasari2020streaming, zhang2020improving}, medical imaging \cite{greenspan2009super, ahmad2022new}, etc. However, in practical situations, acquired videos are often blurred due to camera or object motions \cite{zhang2018adversarial, bahat2017non, zhang2020deblurring}, leading to a deterioration in perceptual quality. Therefore, joint restoration (VSRDB) of VSR and deblurring is needed, which is challenging to achieve the desired level of high-quality videos because two types of degradation in blurry LR videos should be handled simultaneously.

A straightforward approach to solving the joint problem of SR and deblurring is to perform the two tasks sequentially, \textit{i.e.,} by performing SR first and then deblurring, or vice versa. However, this approach has a drawback with the propagation of estimation errors from the preceding operation (SR or deblurring) to the following one (deblurring or SR) \cite{oh2022demfi}. To overcome this, several works proposed joint learning methods of image SR and deblurring (ISRDB), and VSRDB methods \cite{dong2011image, zhang2018gated, xi2021pixel, zhang2020joint, he2015joint, fang2022high}. They showed that the two tasks are strongly inter-correlated. However, most of these methods are designed for ISRDB \cite{dong2011image, zhang2018gated, xi2021pixel, zhang2020joint, he2015joint}. Since motion blur occurs due to camera shakes or object motions, efficient deblurring requires the use of temporal information over video sequences. Recently, Fang \textit{et al.} \cite{fang2022high} proposed the first deep learning-based VSRDB method, called HOFFR, which combines features from the SR and deblurring branches using a parallel fusion module. Although HOFFR exhibited promising performance compared to the ISRDB methods, it struggled to effectively deblur spatially-variant motion blur due to the nature of 2D convolutional neural networks (CNNs) with spatially-equivariant and input-independent filters.

\begin{figure}[t]
\centering
\includegraphics[width=8.3cm]{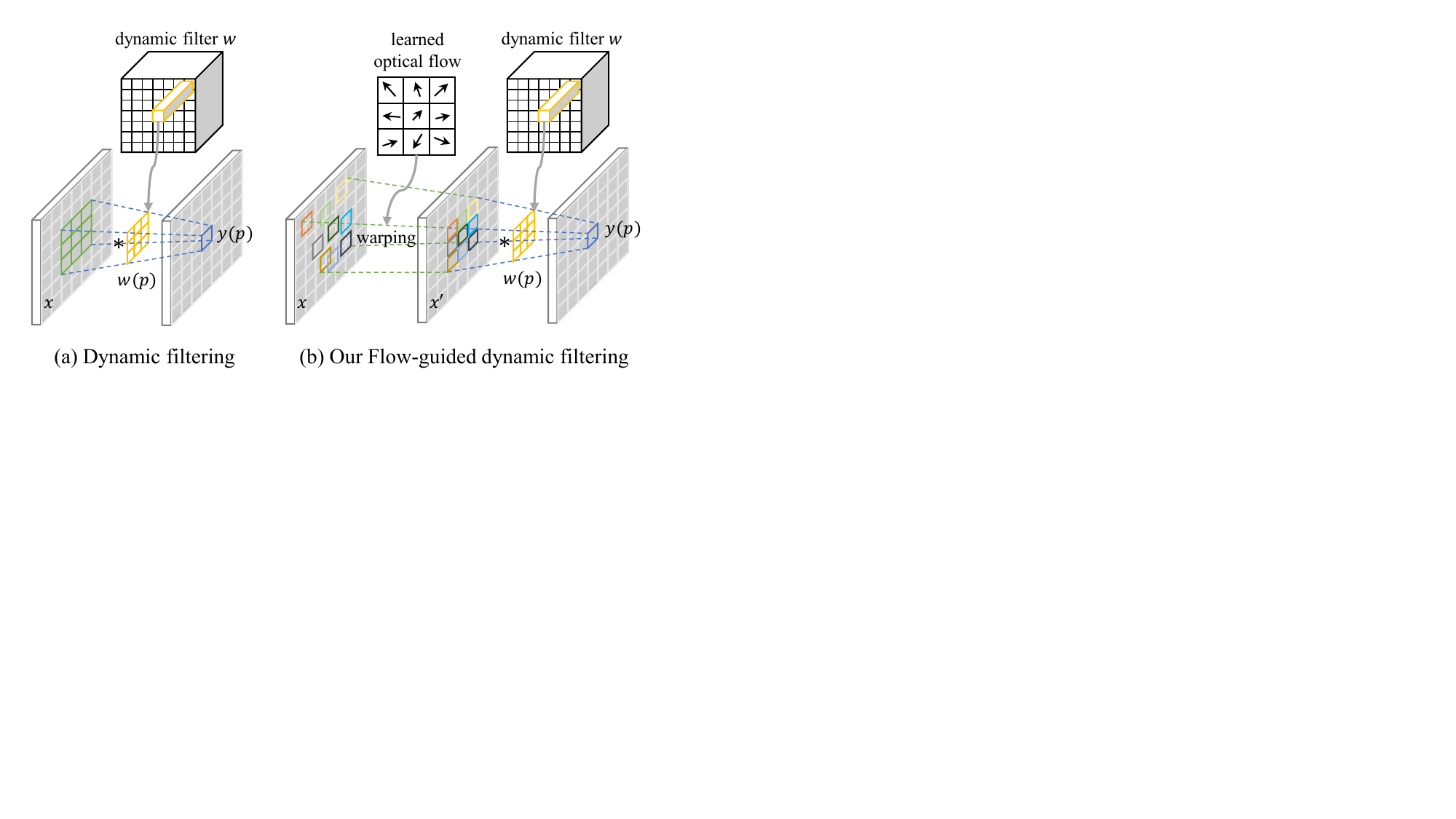}
\caption{Comparison of $3 \times 3$ dynamic filtering. (a) conventional dynamic filtering at location $p$ with fixed surroundings and (b) our flow-guided dynamic filtering (FGDF, Sec. \ref{FGDF}) at position $p$ with variable surroundings guided by learned optical flow.}
\vspace{-5mm}
\label{fig:concept}
\end{figure}

Inspired by the Dynamic Filter Network \cite{jia2016dynamic} in video prediction, significant progress has been made with the dynamic filter mechanism in low-level vision tasks \cite{jo2018deep, niklaus2017video, niklaus2017video2, kim2021koalanet, zhou2019spatio}. Specifically, SR \cite{jo2018deep, kim2021koalanet} and deblurring \cite{zhou2019spatio, fang2023self} have shown remarkable performances with this mechanism in predicting spatially-variant degradation or restoration kernels. For example, Zhou \textit{et al.} \cite{zhou2019spatio} proposed a video deblurring method using spatially adaptive alignment and deblurring filters. However, this method applies filtering only to the reference frame, which limits its ability to accurately exploit information from adjacent frames. To fully utilize motion information from adjacent frames, large-sized filters are required to capture large motions, resulting in high computational complexity. While the method \cite{niklaus2017video2} of using two separable large 1D kernels to approximate a large 2D kernel seems feasible, it loses the ability to capture fine detail, making it difficult to apply for video effectively.

We propose FMA-Net, a novel VSRDB framework based on Flow-Guided Dynamic Filtering (FGDF) and an Iterative Feature Refinement with Multi-Attention (FRMA), to allow for small-to-large motion representation learning with good joint restoration performance. The key insight of the FGDF is to perform filtering that is aware of motion trajectories rather than sticking to fixed positions, enabling effective handling of large motions with small-sized kernels. Fig. \ref{fig:concept} illustrates the concept of our FGDF. The FGDF looks similar to the deformable convolution (DCN) \cite{dai2017deformable} but is different in that it learns position-wise $n \times n$ dynamic filter coefficients, while the DCN learns position-invariant $n \times n$ filter coefficients. 

Our FMA-Net consists of (i) a degradation learning network that estimates motion-aware spatio-temporally-variant degradation kernels and (ii) a restoration network that utilizes these predicted degradation kernels to restore the blurry LR video. The newly proposed multi-attention, consisting of center-oriented attention and degradation-aware attention, enables the FMA-Net to focus on the target frame and utilize the degradation kernels in a globally adaptive manner for VSRDB. We empirically show that the proposed FMA-Net significantly outperforms the recent state-of-the-art (SOTA) methods for video SR and deblurring in objective and subjective qualities on the REDS4, GoPro, and YouTube test datasets under a fair comparison, demonstrating its good generalization ability.


\vspace{-3mm}
\section{Related Work}

\begin{figure*}[t]
\centering
\includegraphics[width=15cm]{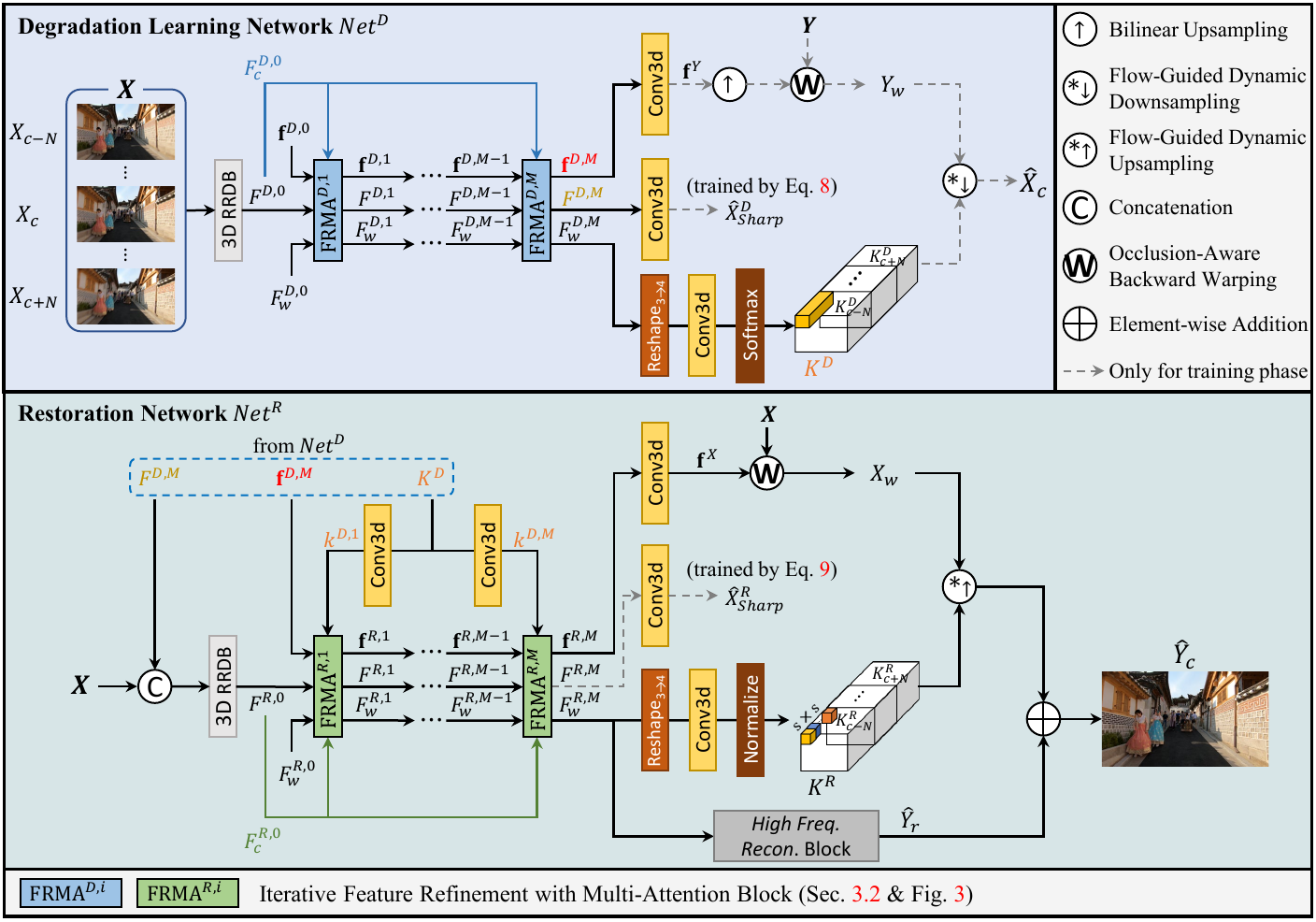}
\caption{The architecture of FMA-Net for video super-resolution and deblurring (VSRDB).}
\vspace{-5mm}
\label{fig:network}
\end{figure*}


\subsection{Video Super-Resolution}
In contrast to image SR that focuses primarily on extracting essential features \cite{dong2014learning, kim2016accurate, zhang2018learning, zhang2018image, kim2021koalanet} and capturing spatial relationships \cite{liang2021swinir, chen2023activating}, VSR faces with an additional key challenge of efficiently utilizing highly correlated but misaligned frames. Based on the number of input frames, VSR is mainly categorized into two types: sliding window-based methods \cite{caballero2017real, huang2017video, jo2018deep, tian2020tdan, wang2020deep, li2020mucan, isobe2020video, li2023simple} and recurrent-based methods \cite{fuoli2019efficient, haris2019recurrent, sajjadi2018frame, lin2021fdan, chan2021basicvsr, chan2022basicvsr++, liu2022learning}.

\noindent \textbf{Sliding window-based methods.}\quad Sliding window-based methods aim to recover HR frames by using neighboring frames within a sliding window. These methods mainly employ CNNs \cite{isobe2020video, jo2018deep, kim20183dsrnet, li2019fast}, optical flow estimation \cite{caballero2017real, tao2017detail}, deformable convolution (DCN) \cite{dai2017deformable, tian2020tdan, wang2019edvr}, or Transformer structures \cite{li2020mucan, cao2021vsrt, liang2022vrt}, with a focus on temporal alignment either explicitly or implicitly. 



\noindent \textbf{Recurrent-based methods.}\quad Recurrent-based methods sequentially propagate the latent features of one frame to the next frame. BasicVSR \cite{chan2021basicvsr} and BasicVSR++ \cite{chan2022basicvsr++} introduced the VSR methods by combining bidirectional propagation of the past and future frames into the features of the current frame, achieving significant improvements. However, the recurrent mechanism is prone to gradient vanishing \cite{hochreiter1998vanishing, chiche2022stable, liu2022learning}, thus causing information loss to some extent. 

Although some progress has been made, all the above methods can handle not blurry but sharp LR videos.


\subsection{Video Deblurring}
Video deblurring aims to remove blur artifacts from blurry input videos. It can be categorized into single-frame deblurring \cite{kupyn2018deblurgan, zamir2022restormer, tsai2022stripformer, kong2023efficient} and multi-frame deblurring \cite{hyun2015generalized, jin2018learning, kim2017dynamic, liang2022vrt, liang2022recurrent}. Zhang \textit{et al.} \cite{zhang2018adversarial} proposed a 3D CNN-based deblurring method to handle spatio-temporal features, while Li \textit{et al.} \cite{li2023simple} introduced a deblurring method based on grouped spatial-temporal shifts. Recently, transformer-based deblurring methods such as Restormer \cite{zamir2022restormer}, Stripformer \cite{tsai2022stripformer}, and RVRT \cite{liang2022recurrent} have been proposed and demonstrated significant performance improvements.

\subsection{Dynamic Filtering-based Restoration}
In contrast to conventional CNNs with spatially-equivariant filters, Jia \textit{et al.} \cite{jia2016dynamic} proposed a dynamic filter network that predicts conditioned kernels for input images and filters the images in a locally adaptive manner. Subsequently, Jo \textit{et al.} \cite{jo2018deep} introduced dynamic upsampling for VSR, while Niklaus \textit{et al.} \cite{niklaus2017video, niklaus2017video2} applied dynamic filtering for frame interpolation. Zhou \textit{et al.} \cite{zhou2019spatio} proposed a spatially adaptive deblurring filter for recurrent video deblurring, and Kim \textit{et al.} \cite{kim2021koalanet} proposed KOALAnet for blind SR, which predicts spatially-variant degradation and upsampling filters. However, all these methods operate naively on a target position and its fixed surrounding neighbors of images or features and cannot effectively handle spatio-temporally-variant motion.

\subsection{Joint Video Super-Resolution and Deblurring}

Despite very active deep learning-based research on single restoration problems such as VSR \cite{chan2021basicvsr, liu2022learning, tian2020tdan, li2020mucan, jo2018deep} and deblurring \cite{liang2022recurrent, liang2022vrt, kim2017dynamic, jin2018learning}, the joint restoration (VSRDB) of these two tasks has drawn much less attention. Recently, Fang \textit{et al.} \cite{fang2022high} introduced HOFFR, the first deep learning-based VSRDB framework. Although they have demonstrated that the HOFFR outperforms ISRDB or sequential cascade approaches of SR and deblurring, the performance has not been significantly elevated, mainly due to the inherent characteristics of 2D CNNs with spatially-equivariant and input-independent filters. Therefore, there still remain many avenues for improvement, especially in effectively restoring spatio-temporally-variant degradations.


\section{Proposed Method}

\subsection{Overview of FMA-Net}
We aim to perform video super-resolution and deblurring (VSRDB) simultaneously. Let a blurry LR input sequence $\bm{X}=\{X_{c-N:c+N}\} \in \mathbb{R}^{T \times H \times W \times 3}$, where $T=2N+1$ and $c$ denote the number of input frames and a center frame index, respectively. Our goal of VSRDB is set to predict a sharp HR center frame $\hat{Y}_c \in \mathbb{R}^{sH \times sW \times 3}$, where $s$ represents the SR scale factor. Fig. \ref{fig:network} illustrates the architecture of our proposed VSRDB framework, FMA-Net. The FMA-Net consists of (i) a degradation learning network $Net^D$ and (ii) a restoration network $Net^R$. $Net^D$ predicts motion-aware spatio-temporally-variant degradation, while $Net^R$ utilizes the predicted degradation from $Net^D$ in a globally adaptive manner to restore the center frame $X_c$. Both $Net^D$ and $Net^R$ have a similar structure, consisting of the proposed iterative feature refinement with multi-attention (FRMA) blocks and a flow-guided dynamic filtering (FGDF) module. Therefore, in this section, we first describe the FRMA block and FGDF in Sec. \ref{FRMA} and Sec. \ref{FGDF}, respectively. Then, we explain the overall structure of FMA-Net in Sec. \ref{arch}. Finally, we present the loss functions and training strategy for the FMA-Net training in Sec. \ref{strategy}.

\subsection{Iterative Feature Refinement with Multi-Attention (FRMA)} \label{FRMA}

We use both types of image-based and feature-based optical flows to capture motion information in blurry videos and leverage them to align and enhance features. However, directly using a pre-trained optical flow network is unstable for blurry frames and computationally expensive \cite{oh2022demfi}. To overcome this instability, we propose the FRMA block. The FRMA block is designed to learn self-induced optical flow and features in a residual learning manner, and we stack $M$ FRMA blocks to iteratively refine features. Notably, inspired by \cite{chan2021understanding, hu2022many}, the FRMA block learns multiple optical flows with their corresponding occlusion masks. This flow diversity enables the learning of one-to-many relations between pixels in a target frame and its neighbor frames, which is beneficial for blurry frames where pixel information is spread due to light accumulation \cite{gupta2010single, harmeling2010space}. 

\begin{figure}
\centering
\includegraphics[width=8cm]{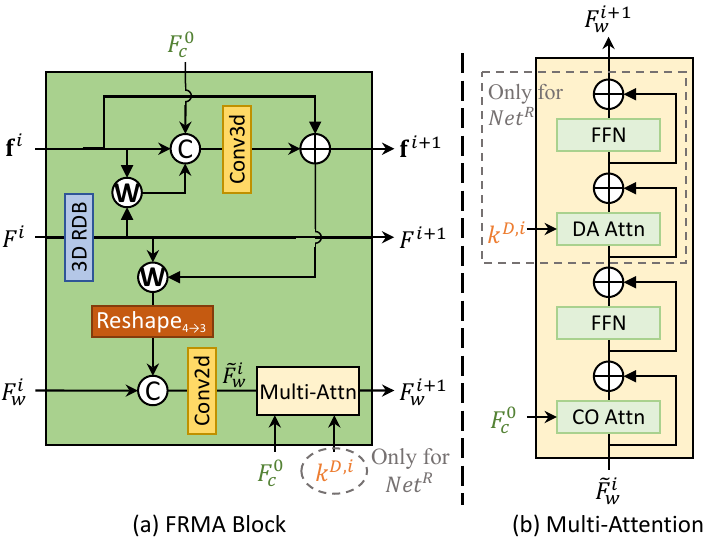}
\vspace{-1mm}
\caption{(a) Structure of \textit{i}$+1$\textit{-th} FRMA block (Sec. \ref{FRMA}); (b) Structure of Multi-Attention. FFN refers to the feed-forward network of the transformer \cite{vaswani2017attention, dosovitskiy2020image}.}
\vspace{-5mm}

\label{fig:FRMA}
\end{figure}

Fig. \ref{fig:FRMA}{\color{red}(a)} illustrates the structure of the FRMA block at the (\textit{i}$+1$)-\textit{th} update-step. Note that FRMA block is incorporated into both $Net^D$ and $Net^R$. To explain the operation of the FRMA block, we omit the superscript $D$ and $R$ for simplicity from its input and output notions in Fig. \ref{fig:network}. The FRMA block aims to refine three tensors: temporally-anchored (unwarped) feature $F \in \mathbb{R}^{T \times H \times W \times C}$ at each frame index, warped feature $F_w \in \mathbb{R}^{H \times W \times C}$, and multi-flow-mask pairs $\textbf{f}\equiv\{f^{j}_{c\rightarrow(c+t)},o^{j}_{c\rightarrow(c+t)}\}_{j=1:n}^{t=-N:N} \in \mathbb{R}^{T \times H \times W \times (2+1)n}$, where $n$ denotes the number of multi-flow-mask pairs from the center frame index $c$ to each frame index, including learnable occlusion masks $o^{j}_{c\rightarrow(c+t)}$ which are sigmoid activations for stability \cite{oh2022demfi}.

\noindent \textbf{(\textit{i}+1)-\textit{th} Feature Refinement.}\quad Given the features $F^i$, $F_w^i$, and $\textbf{f}^i$ computed at the \textit{i}-\textit{th} update-step, we sequentially update each of these features. First, we refine $F^i$ through a 3D RDB \cite{zhang2018residual} to compute $F^{i+1}$ as shown in Fig. \ref{fig:FRMA}{\color{red}(a)}, \textit{i.e.,} $F^{i+1}=\mbox{RDB}(F^i)$. Then, we update $\textbf{f}^i$ to $\textbf{f}^{i+1}$, by warping $F^{i+1}$ to the center frame index $c$ based on $\textbf{f}^i$ and concatenating the resultant with $F_c^0$ and $\textbf{f}^i$, which is given as:
\vspace{-1mm}
\begin{equation} \label{eq:flow_update}
    \textbf{f}^{i+1} = \textbf{f}^i + \mbox{Conv}_{3d}(\mbox{concat}(\textbf{f}^i, \mathcal{W}(F^{i+1}, \textbf{f}^i), F_c^0)),
\end{equation}
\noindent where $\mathcal{W}$ and $\mbox{concat}$ denote the occlusion-aware backward warping \cite{jaderberg2015spatial, oh2022demfi, sim2021xvfi} and concatenation along channel dimension, respectively. Note that $F_c^0 \in \mathbb{R}^{H \times W \times C}$ represents the feature map at the center frame index $c$ of the initial feature $F^0 \in \mathbb{R}^{T \times H \times W \times C}$. Finally, we update $F_w^i$ by using warped $F^{i+1}$ to the center frame index $c$ by $\textbf{f}^{i+1}$ as:
\vspace{-1mm}
\begin{equation} \label{eq:WF_update}
    \Tilde{F}_w^i = \mbox{Conv}_{2d}(\mbox{concat}(F_w^i, r_{4 \rightarrow 3}(\mathcal{W}(F^{i+1}, \textbf{f}^{i+1})))),
\end{equation}
\noindent where $r_{4 \rightarrow 3}$ denotes the reshape operation from $\mathbb{R}^{T \times H \times W \times C}$ to $\mathbb{R}^{H \times W \times TC}$ for feature aggregation.

\noindent \textbf{Multi-Attention.}\quad Our multi-attention structure is shown in the Fig. \ref{fig:FRMA}{\color{red}(b)}. To better align $\Tilde{F}_w^i$ to the center frame index $c$ and adapt to spatio-temporally variant degradation, we enhance $\Tilde{F}_w^i$ using center-oriented (CO) attention and degradation-aware (DA) attention. In the case of `CO attention', for the input $\Tilde{F}_w^i$ and $F_c^0$, it generates \textit{query} ($Q$), \textit{key} ($K$), and \textit{value} ($V$) as $Q = W_qF_c^0$, $K = W_k\Tilde{F}_w^i$, and $V = W_v\Tilde{F}_w^i$, respectively. Then, we calculate the attention map between $Q$ and $K$, and use it to adjust $V$. While this process may resemble self-attention \cite{dosovitskiy2020image, vaswani2017attention} at first, our empirical findings indicate better performance when $\Tilde{F}_w^i$ focuses on its relation with $F_c^0$ rather than on itself. The CO attention process is expressed as:
\vspace{-1mm}
\begin{equation} \label{eq:attention}
    \mbox{CO Attention}(Q,K,V) = \mbox{SoftMax}(QK^T/\sqrt{d})V,
\end{equation}
\noindent where $\sqrt{d}$ denotes the scaling factor \cite{dosovitskiy2020image, vaswani2017attention}. The `DA attention' is the same as the CO attention except that the \textit{query} is derived from feature $k^{D,i} \in \mathbb{R}^{H \times W \times C}$, which is adjusted by convolution with the novel motion-aware degradation kernels $K^D$ from $Net^D$, rather than from $F_c^0$. This process enables $\Tilde{F}_w^i$ to be globally adaptive to degradation. The motion-aware kernel $K^D$ will be described in detail in Sec. \ref{arch}. It should be noted that DA attention is only applied in $Net^R$ since it utilizes the predicted $K^D$ from $Net^D$ as shown in Fig. \ref{fig:FRMA}. Specifically, we empirically found out that the adoption of the transposed-attention \cite{zamir2022restormer, ali2021xcit} in Eq. \ref{eq:attention} shows more efficient and better performances.


\subsection{Flow-Guided Dynamic Filtering} \label{FGDF}

We start with a brief overview of dynamic filtering \cite{jia2016dynamic}. Let $p_k$ represent the $k$\textit{-th} sampling offset in a standard convolution with a kernel size of $n \times n$. For instance, we have $p_k \in \{(-1,-1),(-1,0),\dots,(1,1)\}$ when $n=3$. We denote the predicted $n \times n$ dynamic filter at position $p$ as $F^p$. The dynamic filtering for images can be formulated as:
\vspace{-2mm}
\begin{equation} \label{eq:dynamic_filtering}
    y(p) = \sum_{k=1}^{n^2} F^p(p_k) \cdot x(p+p_k),
\end{equation}
\noindent where $x$ and $y$ are the input and output features. Its naive extension to video can be expressed as:
\vspace{-2mm}
\begin{equation} \label{eq:video_dynamic_filtering}
    y(p) = \sum_{t=-N}^{+N}\sum_{k=1}^{n^2} F^p_{c+t}(p_k) \cdot x_{c+t}(p+p_k),
\end{equation}
\noindent where $c$ represents the center frame index of the input frames. However, such a naive extension of filtering at a pixel position with fixed surrounding neighbors requires a large-sized filter to capture large motions, resulting in an exponential increase in computation and memory usage. To overcome this problem, we propose flow-guided dynamic filtering (FGDF) inspired by DCN \cite{dai2017deformable}. The kernels are dynamically generated to be pixel-wise motion-aware, guided by the optical flow. This allows effective handling of large motion with relatively small-sized kernels. Our FGDF can be formulated as:
\vspace{-2mm}
\begin{equation} \label{eq:flow_guided_dynamic_filtering}
    \begin{aligned}
        y(p) &= \sum_{t=-N}^{+N}\sum_{k=1}^{n^2} F^p_{c+t}(p_k) \cdot x'_{c+t}(p+p_k),
    \end{aligned}
\end{equation}
\noindent where $x'_{c+t} = \mathcal{W}(x_{c+t}, \textbf{f}_{c+t})$ and $\textbf{f}_{c+t}$ denotes optical flow with its occlusion mask from frame index $c$ to $c+t$.

\subsection{Overall Architecture} \label{arch}

\noindent \textbf{Degradation Learning Network.} \quad $Net^D$, shown in the upper part of Fig. \ref{fig:network}, takes a blurry LR sequence $\bm{X}$ as input and aims to predict a motion-aware spatio-temporally variant degradation kernels that are assumed to be used to obtain center frame $X_c$ from the sharp HR counterpart $\bm{Y}$. Specifically, we first compute the initial temporally-anchored feature $F^{D,0}$ from $\bm{X}$ through a 3D RRDB \cite{wang2018esrgan}. Then, we refine $F^{D,0}$, ${F}_w^{D,0}$, and $\textbf{f}^{D,0}$ through $M$ FRMA blocks (Eqs. \ref{eq:flow_update} and \ref{eq:WF_update}). Meanwhile, ${F}_w^{D,i}$ is adaptively adjusted in the CO attention of each FRMA block based on its relation to $F_c^{D,0}$ (Eq. \ref{eq:attention}), the \textit{center} feature map of $F^{D,0}$. It should be noted that we initially set ${F}_w^{D,0}=\textbf{0}$ and $\textbf{f}^{D,0}=\{f^{j}_{c\rightarrow(c+t)}=\textbf{0},o^{j}_{c\rightarrow(c+t)}=\textbf{1}\}_{j=1:n}^{t=-N:N}$. Subsequently, using the final refined features $\textbf{f}^{D,M}$ and ${F}_w^{D,M}$, we calculate an \textit{image} flow-mask pair $\textbf{f}^Y \in \mathbb{R}^{T \times H \times W \times (2+1)}$ for $Y$ and its corresponding motion-aware degradation kernels $K^D \in \mathbb{R}^{T \times H \times W \times k_d^2}$, where $k_d$ denotes the degradation kernel size. Here, we use a sigmoid function to normalize $K^D$, which mimics the blur generation process \cite{oh2022demfi, nah2017deep, su2017deep, shen2020blurry} where all kernels have positive values. Finally, we synthesize $\hat{X}_c$ with $K^D$ and $\textbf{f}^Y$ as:
\begin{equation} \label{eq:recon}
    \hat{X}_c = (\mathcal{W}(\bm{Y}, s \cdot (\textbf{f}^Y\uparrow_s)) \oast K^D)\downarrow_s,
\end{equation}
\noindent where $\oast\downarrow_s$ represents novel $k_d \times k_d$ FGDF via Eq. \ref{eq:flow_guided_dynamic_filtering} at each pixel location with stride $s$ and $\uparrow_s$ denotes $\times s$ bilinear upsampling. Additionally, $F^{D,M}$ is mapped to the image domain via 3D convolution to generate $\hat{X}_{Sharp}^D \in \mathbb{R}^{T \times H \times W \times 3}$, which is only used to train the network.

\noindent \textbf{Restoration Network.}\quad $Net^R$ differs from $Net^D$ which predicts flow and degradation in $\bm{Y}$. Instead, $Net^R$ computes the flow in $\bm{X}$ and utilizes it along with the predicted $K^D$ for VSRDB. $Net^R$ takes $\bm{X}$, $F^{D,M}$, $\textbf{f}^{D,M}$, and $K^D$ as inputs. It first computes $F^{R,0}$ through a concatenation of $\bm{X}$ and $F^{D,M}$ using a RRDB and then refines three features, $F^{R,0}$, ${F}_w^{R,0}$, and $\textbf{f}^{R,0}$ through the cascaded $M$ FRMA blocks. Notably, we set ${F}_w^{R,0}=\textbf{0}$ and $\textbf{f}^{R,0}=\textbf{f}^{D,M}$ in this case. During this FRMA process, each ${F}_w^{R,i}$ is globally adjusted based on both $F_c^{R,0}$ and the adjusted kernel $k^{D,i}$ through CO and DA attentions, where $k^{D,i}$ represents the degradation features adjusted by convolutions from $K^D$. Subsequently, $\textbf{f}^{R,M}$ is used to generate an image flow-mask pair $\textbf{f}^X \in \mathbb{R}^{T \times H \times W \times (2+1)}$ for $\bm{X}$, while ${F}_w^{R,M}$ is used to generate the high-frequency detail $\hat{Y}_r$ and the pixel-wise motion-aware $\times s$ upsampling and deblurring (\textit{i.e.} restoration) kernels $K^R \in \mathbb{R}^{T \times H \times W \times s^2k_r^2}$ for warped $\bm{X}$, where $k_r$ denotes the restoration kernel size. $\hat{Y}_r$ is generated by stacked convolution and pixel shuffle \cite{shi2016real} (\textit{High Freq. Recon.} Block in Fig. \ref{fig:network}). The pixel-wise kernels $K^R$ are normalized with respect to all kernels at temporally co-located positions over $\bm{X}$, similar to \cite{kim2021koalanet}. Finally, $\hat{Y}_c$ can be obtained as $\hat{Y}_c = \hat{Y}_r + (\mathcal{W}(\bm{X}, \textbf{f}^X) \oast K^R)\uparrow_s$, where $\oast\uparrow_s$ represents proposed flow-guided $\times s$ dynamic upsampling at each pixel location based on Eq. \ref{eq:flow_guided_dynamic_filtering}. Furthermore, $F^{D,R}$ is also mapped to the image domain through 3D convolution, similar to $Net^D$, to generate $\hat{X}_{Sharp}^R \in \mathbb{R}^{T \times H \times W \times 3}$, which is only used in FMA-Net training.




\renewcommand{\arraystretch}{1.0}
\begin{table*}[t]
\begin{center}
\setlength\tabcolsep{20pt} 
\scalebox{0.75}{
\begin{tabular}{cccc}
\bottomrule
\multicolumn{1}{c|}{\multirow{2}{*}{Methods}} & \multicolumn{1}{c|}{\multirow{2}{*}{\# Params (M)}} & \multicolumn{1}{c|}{\multirow{2}{*}{Runtime (s)}} & REDS4             \\
\multicolumn{1}{c|}{}                         & \multicolumn{1}{c|}{}                               & \multicolumn{1}{c|}{}                             & PSNR $\uparrow$ / SSIM $\uparrow$ / tOF $\downarrow$ \\ 
\hline
\multicolumn{4}{c}{Super-Resolution + Deblurring}                                                                                                                                                                         \\ 
\hline
\multicolumn{1}{c|}{SwinIR \cite{liang2021swinir} + Restormer \cite{zamir2022restormer}}       & \multicolumn{1}{c|}{11.9 + 26.1}                              & \multicolumn{1}{c|}{0.320 + 1.121}                            & 24.33 / 0.7040 / 4.82                 \\
\multicolumn{1}{c|}{HAT \cite{chen2023activating} + FFTformer \cite{kong2023efficient}}          & \multicolumn{1}{c|}{20.8 + 16.6}                              & \multicolumn{1}{c|}{0.609 + 1.788}                            & 24.22 / 0.7091 / 4.40                 \\
\multicolumn{1}{c|}{BasicVSR++ \cite{chan2022basicvsr++} + RVRT \cite{liang2022recurrent}}        & \multicolumn{1}{c|}{7.3 + 13.6}                              & \multicolumn{1}{c|}{0.072 + 0.623}                            & 24.92 / 0.7604 / 3.49                 \\
\multicolumn{1}{c|}{FTVSR \cite{qiu2022learning} + GShiftNet \cite{li2023simple}}        & \multicolumn{1}{c|}{45.8 + 13.0}                              & \multicolumn{1}{c|}{0.527 + 2251}                            & 24.72 / 0.7415 / 3.69                 \\ 
\hline
\multicolumn{4}{c}{Deblurring + Super-Resolution}                                                                                                                                                                         \\ 
\hline
\multicolumn{1}{c|}{Restormer \cite{zamir2022restormer} + SwinIR \cite{liang2021swinir}}       & \multicolumn{1}{c|}{26.1 + 11.9}                              & \multicolumn{1}{c|}{0.078 + 0.320}                            & 24.30 / 0.7085 / 4.49                 \\
\multicolumn{1}{c|}{FFTformer \cite{kong2023efficient} + HAT \cite{chen2023activating}}          & \multicolumn{1}{c|}{16.6 + 20.8}                              & \multicolumn{1}{c|}{0.124 + 0.609}                            & 24.21 / 0.7111 / 4.38                 \\
\multicolumn{1}{c|}{RVRT \cite{liang2022recurrent} + BasicVSR++ \cite{chan2022basicvsr++}}        & \multicolumn{1}{c|}{13.6 + 7.3}                              & \multicolumn{1}{c|}{0.028 + 0.072}                            & 24.79 / 0.7361 / 3.66                 \\
\multicolumn{1}{c|}{GShiftNet \cite{li2023simple} + FTVSR \cite{qiu2022learning}}        & \multicolumn{1}{c|}{13.0 + 45.8}                              & \multicolumn{1}{c|}{0.102 + 0.527}                            & 23.47 / 0.7044 / 3.98                 \\ 
\hline
\multicolumn{4}{c}{Joint Video Super-Resolution and Deblurring}                                                                                                                                                                                             \\ 
\hline
\multicolumn{1}{c|}{HOFFR \cite{fang2022high}}                    & \multicolumn{1}{c|}{3.5}                              & \multicolumn{1}{c|}{0.500}                            & 27.24 / 0.7870 / -                 \\
\multicolumn{1}{c|}{Restormer$^*$ \cite{zamir2022restormer}}               & \multicolumn{1}{c|}{26.5}                              & \multicolumn{1}{c|}{0.081}                            & 27.29 / 0.7850 / 2.71                 \\
\multicolumn{1}{c|}{GShiftNet$^*$ \cite{li2023simple}}               & \multicolumn{1}{c|}{13.5}                              & \multicolumn{1}{c|}{0.185}                            & 25.77 / 0.7275 / 2.96                 \\ 
\multicolumn{1}{c|}{BasicVSR++$^*$ \cite{chan2022basicvsr++}}               & \multicolumn{1}{c|}{7.3}                              & \multicolumn{1}{c|}{0.072}                            & 27.06 / 0.7752 / 2.70                 \\
\multicolumn{1}{c|}{RVRT$^*$ \cite{liang2022recurrent}}                    & \multicolumn{1}{c|}{12.9}                              & \multicolumn{1}{c|}{0.680}                            & {\color{blue}27.80} / {\color{blue}0.8025} / {\color{blue}2.40}                 \\
\hline
\multicolumn{1}{c|}{FMA-Net (Ours)}                     & \multicolumn{1}{c|}{9.6}                              & \multicolumn{1}{c|}{0.427}                            & {\color{red}\textbf{28.83}} / {\color{red}\textbf{0.8315}} / {\color{red}\textbf{1.92}}          \\ 
\toprule
\end{tabular}}
\end{center}
\vspace{-5mm}
\caption{Quantitative comparison on REDS4 for $\times 4$ VSRDB. All results are calculated on the RGB channel. {\color{red}\textbf{Red}} and {\color{blue}blue} colors indicate the best and second-best performance, respectively. Runtime is calculated on an LR frame sequence of size $180 \times 320$. The superscript $^*$ indicates that the model is retrained on the REDS \cite{nah2019ntire} training dataset for VSRDB.}
\vspace{-5mm}
\label{tab:reds4_comparison}
\end{table*}

\subsection{Training Strategy} \label{strategy}
We employ a two-stage training strategy to train the FMA-Net. $Net^D$ is first pre-trained with the loss $L_D$ as:
\vspace{-5mm}
\begin{align} \label{eq:loss_d}
    L_D &= l_1(\hat{X}_c,X_c) + \lambda_1 \sum_{t=-N}^{+N}l_1(\mathcal{W}(Y_{t+c}, s \cdot (\textbf{f}^Y_{t+c}\uparrow_s)),Y_c) \notag \\
    \begin{split}
        + \lambda_2 l_1(f^Y, f^Y_{RAFT}) + \lambda_3 \underbrace{l_1(\hat{X}_{Sharp}^D, X_{Sharp})}_{\text{Temporal Anchor (TA) loss}},
    \end{split}
\end{align}

\vspace{-4mm}

\noindent where $f^Y$ represents the image optical flow contained in $\textbf{f}^Y$, and $f^Y_{RAFT}$ denotes the pseudo-GT optical flow generated by a pre-trained RAFT \cite{teed2020raft} model. $X_{Sharp}$ is the sharp LR sequence obtained by applying bicubic downsampling to $\bm{Y}$. The first term on the right side in Eq. \ref{eq:loss_d} is the reconstruction loss, the second term is the warping loss for optical flow learning in $\bm{Y}$ from center frame index $c$ to $c+t$, and the third term is the loss using RAFT pseudo-GT for further refining the optical flow.\\
\textbf{Temporal Anchor (TA) Loss.} \quad 
Finally, to boost performance, we propose a TA loss, the last term on the right side in Eq. \ref{eq:loss_d}. This loss sharpens $F^D$ while keeping each feature temporally anchored for the corresponding frame index, thus constraining the solution space according to our intention to distinguish warped and unwarped features. 

After pre-training, the FMA-Net in Fig. \ref{fig:network} is jointly trained as the second stage training with the total loss $L_{total}$:
\vspace{-6mm}
\begin{align} \label{eq:loss_total}
    L_{total} &= l_1(\hat{Y}_c,Y_c) + \lambda_4 \sum_{t=-N}^{+N}l_1(\mathcal{W}(X_{t+c}, \textbf{f}^X_{t+c}),X_c) \notag \\
    \begin{split}
        &+\lambda_5 \underbrace{l_1(\hat{X}_{Sharp}^R, X_{Sharp})}_{\text{Temporal Anchor (TA) loss}} + \lambda_6 L_D,
    \end{split}
\end{align}

\vspace{-3mm}

\noindent where the first term on the right side is the restoration loss, and the second and third terms are identical to the second and forth terms in Eq. \ref{eq:loss_d}, except for their applied domains.

\section{Experiment Results}

\noindent \textbf{Implementation details.}\quad We train the FMA-Net using the Adam optimizer \cite{kingma2014adam} with a mini-batch size of $8$. The initial learning rate is set to $2 \times 10^{-4}$, and reduced by half at $70\%$, $85\%$, and $95\%$ of total $300K$ iterations in each training stage. The training LR patch size is $64 \times 64$, the number of FRMA blocks is $M=4$, the number of multi-flow-mask pairs is $n=9$, and the kernel sizes $k_d$ and $k_r$ are $20$ and $5$, respectively. The coefficients $[\lambda_i]^6_{i=1}$ in Eqs. \ref{eq:loss_d} and \ref{eq:loss_total} are determined through grid searching, with $\lambda_2$ set to $10^{-4}$ and all other values set to $10^{-1}$. We consider $T=3$ (that is, $N=1$) and $s=4$ in our experiments. Additionally, we adopted the multi-Dconv head transposed attention (MDTA) and Gated-Dconv feed-forward network (GDFN) modules proposed in Restormer \cite{zamir2022restormer} for the attention and feed-forward network in our multi-attention block.

\noindent \textbf{Datasets.}\quad We train FMA-Net using the REDS \cite{nah2019ntire} dataset which consists of realistic and dynamic scenes. Following previous works \cite{liu2022learning, li2020mucan, wang2019edvr}, we use REDS4 \footnote{Clips 000, 011, 015, 020 of the REDS training set.} as the test set, while the remaining clips are for training. Also, to evaluate generalization performance, we employ the GoPro \cite{nah2017deep} and YouTube datasets as test sets alongside REDS4. For the GoPro dataset, we applied bicubic downsampling to its blurry version to evaluate VSRDB. As for the YouTube dataset, we selected $40$ YouTube videos of different scenes with a resolution of $720 \times 1,280$ at $240$fps, including extreme scenes from various devices. Subsequently, we temporally and spatially downsampled them, similar to previous works \cite{oh2022demfi, shen2020blurry, gupta2020alanet}, resulting in blurry $30$ fps of $180 \times 320$ size.

\noindent \textbf{Evaluation metrics.}\quad
We use PSNR and SSIM \cite{wang2004image} to evaluate the quality of images generated by the networks, and tOF \cite{chu2020learning, oh2022demfi} to evaluate temporal consistency. We also compare the model sizes and runtime.

\subsection{Comparisons with State-of-the-Art Methods}
To achieve VSRDB, we compare our FMA-Net with the very recent SOTA methods: two single-image SR models (SwinIR \cite{liang2021swinir} and HAT \cite{chen2023activating}), two single-image deblurring models (Restormer \cite{zamir2022restormer} and FFTformer \cite{kong2023efficient}), two VSR models (BasicVSR++ \cite{chan2021basicvsr} and FTVSR \cite{qiu2022learning}), two video deblurring models (RVRT \cite{liang2022recurrent} and GShiftNet \cite{li2023simple}), and one VSRDB model (HOFFR \cite{fang2022high}). Also, we retrain one single-image model (Restormer$^*$ \cite{zamir2022restormer}) and three video models (BasicVSR++$^*$ \cite{chan2022basicvsr++}, GShiftNet$^*$ \cite{li2023simple}, and RVRT$^*$ \cite{liang2022recurrent}) using our training dataset to perform VSRDB for a fair comparison. It should be noted that Restormer$^*$ \cite{zamir2022restormer} is modified to receive concatenated $T$ frames in the channel dimension for video processing instead of a single frame, and we added a pixel-shuffle \cite{shi2016real} block at the end to enable SR.

Table \ref{tab:reds4_comparison} shows the quantitative comparisons for the test set, REDS4. It can be observed in Table \ref{tab:reds4_comparison} that: (i) the sequential approaches of cascading SR and deblurring result in error propagation from previous models, leading to a significant performance drop, and the use of two models also increase memory and runtime costs; (ii) the VSRDB methods consistently demonstrate superior overall performance compared to the sequential cascade approaches, indicating that the two tasks are highly inter-correlated; and (iii) our FMA-Net \textit{significantly} outperforms all SOTA methods including five joint VSRDB methods in terms of PSNR, SSIM, and tOF. Specifically, our FMA-Net achieves improvements of $1.03$ dB and $1.77$ dB over the SOTA algorithms, RVRT$^*$ \cite{liang2022recurrent} and BasicVSR++$^*$ \cite{chan2022basicvsr++}, respectively. The clip-by-clip analyses for REDS4 and the results of all possible combinations of the sequential cascade approaches can be found in the \textit{Supplemental}, including demo videos.


\renewcommand{\arraystretch}{1.1}
\begin{table}[h]
\begin{center}
\scalebox{0.7}{
\begin{tabular}{c|c|c}
\bottomrule
\multirow{2}{*}{Methods} & GoPro             & YouTube           \\
                         & PSNR $\uparrow$ / SSIM $\uparrow$ / tOF $\downarrow$ & PSNR $\uparrow$ / SSIM $\uparrow$ / tOF $\downarrow$ \\ 
\hline
Restormer$^*$ \cite{zamir2022restormer}                &  {\color{blue}26.29} / {\color{blue}0.8278} / 3.66 &  23.94 / 0.7682 / 2.87                \\
GShiftNet$^*$ \cite{li2023simple}                &     25.37 / 0.7922 / 3.95              &  {\color{blue}24.44} / {\color{blue}0.7683} / 2.96                 \\ 
BasicVSR++$^*$ \cite{chan2022basicvsr++}               &    25.19 / 0.7968 / 4.04               &  23.84 / 0.7660 / 2.97                \\
RVRT$^*$ \cite{liang2022recurrent}                    &  25.99 / 0.8267 / {\color{blue}3.55}                 &  23.53 / 0.7588 / {\color{blue}2.78}                 \\
\hline
FMA-Net (Ours)                     &    {\color{red}\textbf{27.65}} / {\color{red}\textbf{0.8542}} / {\color{red}\textbf{3.31}}               &    {\color{red}\textbf{26.02}} / {\color{red}\textbf{0.8067}} / {\color{red}\textbf{2.63}}               \\ 
\toprule

\end{tabular}}
\end{center}
\vspace{-5mm}
\caption{Quantitative comparison on GoPro \cite{nah2017deep} and YouTube test sets for $\times 4$ VSRDB.}
\vspace{-5mm}
\label{tab:gopro_comparison}
\end{table}

\begin{figure}[t]
\centering
\includegraphics[width=8.3cm]{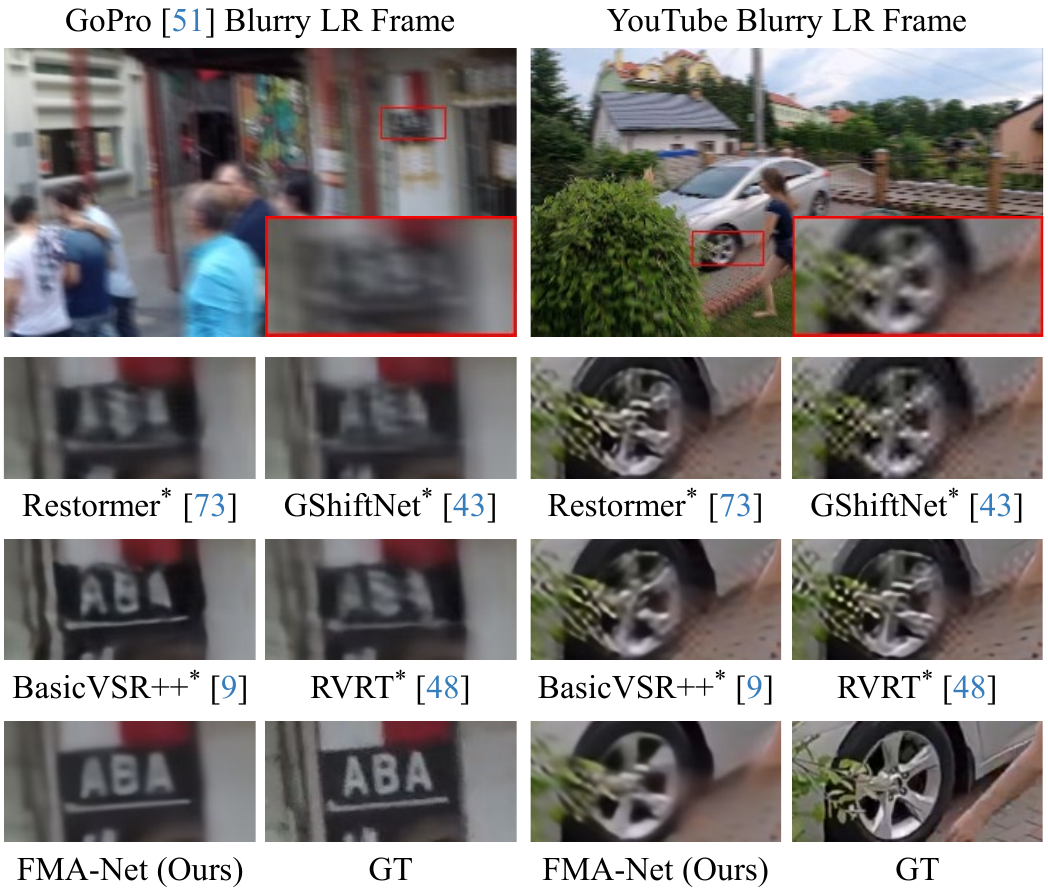}
\caption{Visual results of different methods on REDS4 \cite{nah2019ntire}, GoPro \cite{nah2017deep}, and YouTube test sets. \textit{Best viewed in zoom.}}
\vspace{-5mm}
\label{fig:qualitative_comparison}
\end{figure}

Table \ref{tab:gopro_comparison} shows the quantitative comparisons on GoPro \cite{nah2017deep} and YouTube test sets for \textit{joint} models trained on REDS \cite{nah2019ntire}. When averaged across both test sets, our FMA-Net achieves a performance boost of $2.08$ dB and $1.93$ dB over RVRT$^*$ \cite{liang2022recurrent} and GShiftNet$^*$ \cite{li2023simple}, respectively. This demonstrates that our FMA-Net has good generalization in addressing spatio-temporal degradation generated from various scenes across diverse devices. Figs. \ref{fig:teaser}{\color{red}(a)} and \ref{fig:qualitative_comparison} show the visual results on three test sets, showing that the images generated by our FMA-Net are visually sharper than those by other methods.

\subsection{Ablation Studies}
We analyze the effectiveness of the components in our FMA-Net through ablation studies for which we train the models on REDS \cite{nah2019ntire} and test them on REDS4.

\renewcommand{\arraystretch}{1.1}
\begin{table}[b]
\vspace{-5mm}
\begin{center}
\scalebox{0.68}{
\begin{tabular}{c|c|c|cccc}
\bottomrule
\multirow{2}{*}{$k_d$}  & \multirow{2}{*}{$\textbf{f}$} & \multirow{2}{*}{Network} & \multicolumn{4}{c}{Average Motion Magnitude}                  \\ \cline{4-7} 
                    &                    &                          & $[0, 20)$         & $[20, 40)$        & $\ge 40$            & Total         \\ \hline
\multirow{4}{*}{10} & \multirow{2}{*}{\xmark} & $Net^D$                     & 44.97 / 0.055 & 39.81 / 0.245 & 32.04 / 0.871 & 43.14 / 0.128 \\
                    &                    & $Net^R$                     & 27.85 / 1.713 & 27.51 / 3.922 & 24.69 / 6.857 & 27.69 / 2.489 \\ \cline{2-7} 
                    & \multirow{2}{*}{\cmark} & $Net^D$                     & 45.38 / 0.049 & 42.18 / 0.165 & 37.72 / 0.474 & 44.25 / 0.092 \\
                    &                    & $Net^R$                     & 28.64 / 1.436 & 28.46 / 3.469 & 25.54 / 6.558 & 28.52 / 2.157 \\ \hline
\multirow{4}{*}{20} & \multirow{2}{*}{\xmark} & $Net^D$                     & 45.94 / 0.047 & 42.02 / 0.193 & 35.50 / 0.689 & 44.53 / 0.104 \\
                    &                    & $Net^R$                     & 28.10 / 1.566 & 27.54 / 3.835 & 24.24 / 6.989 & 27.86 / 2.365 \\ \cline{2-7} 
                    & \multirow{2}{*}{\cmark} & $Net^D$                     & {\color{blue}46.57} / {\color{blue}0.041} & {\color{blue}43.49} / {\color{blue}0.151} & {\color{blue}38.23} / {\color{blue}0.430}   & {\color{blue}45.46} / {\color{blue}0.082} \\
                    &                    & $Net^R$                     & {\color{blue}28.91} / {\color{blue}1.289} & {\color{blue}28.91} / {\color{blue}3.057} & {\color{blue}26.17} / {\color{blue}5.841}   & {\color{blue}28.83} / {\color{blue}1.918} \\ \hline
\multirow{4}{*}{30} & \multirow{2}{*}{\xmark} & $Net^D$                     & 46.25 / 0.042 & 42.95 / 0.161 & 37.53 / 0.464 & 45.07 / 0.087 \\
                    &                    & $Net^R$                     & 28.30 / 1.589 & 28.10 / 3.589 & 25.58 / 6.258 & 28.19 / 2.292 \\ \cline{2-7} 
                    & \multirow{2}{*}{\cmark} & $Net^D$                     & {\color{red}\textbf{46.89}} / {\color{red}\textbf{0.037}} & {\color{red}\textbf{44.12}} / {\color{red}\textbf{0.133}} & {\color{red}\textbf{39.30}} / {\color{red}\textbf{0.349}}   & {\color{red}\textbf{45.90}} / {\color{red}\textbf{0.072}} \\
                    &                    & $Net^R$                     & {\color{red}\textbf{28.91}} / {\color{red}\textbf{1.283}} & {\color{red}\textbf{28.98}} / {\color{red}\textbf{3.013}} & {\color{red}\textbf{26.37}} / {\color{red}\textbf{5.666}}   & {\color{red}\textbf{28.89}} / {\color{red}\textbf{1.897}} \\ 
\toprule
\end{tabular}}
\end{center}
\vspace{-5mm}
\caption{Ablation study on the FGDF (PSNR$\uparrow$ / tOF$\downarrow$).}
\vspace{-5mm}
\label{tab:abla_fgdf}
\end{table}

\noindent \textbf{Effect of flow-guided dynamic filtering (FGDF).}\quad Table \ref{tab:abla_fgdf} shows the performance of $Net^D$ and $Net^R$ based on the degradation kernel size $k_d$ and two dynamic filtering methods: the conventional dynamic filtering in Eq. \ref{eq:video_dynamic_filtering} and the FGDF in Eq. \ref{eq:flow_guided_dynamic_filtering}. Average motion magnitude refers to the average absolute optical flow \cite{teed2020raft} magnitude between the two consecutive frames. Table \ref{tab:abla_fgdf} reveals the following observations: (i) conventional dynamic filtering \cite{jia2016dynamic, jo2018deep, kim2021koalanet} is not effective in handling large motion, resulting in a significant performance drop as the degree of motion magnitude increases; (ii) our proposed FGDF demonstrates better reconstruction and restoration performance than the conventional dynamic filtering for all ranges of motion magnitudes. This performance difference becomes more pronounced as the degree of motion magnitude increases. For $k_d=20$, when the average motion magnitude is above 40, the proposed FGDF achieves a restoration performance improvement of $1.93$ dB compared to the conventional method. Additional analysis for Table \ref{tab:abla_fgdf} can be found in the \textit{Supplemental}.

\renewcommand{\arraystretch}{1.1}
\begin{table}[t]
\begin{center}
\scalebox{0.7}{
\begin{tabular}{lccc}
\bottomrule
\multicolumn{1}{c|}{\multirow{2}{*}{Methods}}  & \multicolumn{1}{c|}{\# Params} & \multicolumn{1}{c|}{Runtime} & $Net^R$ (sharp HR $\hat{Y}_c$)                  \\
\multicolumn{1}{c|}{}                         & \multicolumn{1}{c|}{(M)}       & \multicolumn{1}{c|}{(s)}     & PSNR $\uparrow$ / SSIM $\uparrow$ / tOF $\downarrow$      \\ \hline
\multicolumn{4}{c}{The number of multi-flow-mask pairs $n$}                    \\ \hline
\multicolumn{1}{l|}{{\color{red}(a)} $n=1$}                      & \multicolumn{1}{c|}{9.15}      & \multicolumn{1}{c|}{0.424}   & 28.24 / 0.8151 / 2.224 \\
\multicolumn{1}{l|}{{\color{red}(b)} $n=5$}                      & \multicolumn{1}{c|}{9.29}      & \multicolumn{1}{c|}{0.429}  & 28.60 / 0.8258 / 2.054 \\ \hline
\multicolumn{4}{c}{Deformable Convolution \cite{dai2017deformable}}                                                                                                                                           \\ \hline
\multicolumn{1}{l|}{{\color{red}(c)} w/ DCN (\#$\mbox{offset}=9$)}         & \multicolumn{1}{c|}{10.13}     & \multicolumn{1}{c|}{0.426}   & 28.52 / 0.8225 / 2.058 \\ \hline
\multicolumn{4}{c}{Loss Function and Training Strategy}                                                                                                                                                    \\ \hline
\multicolumn{1}{l|}{{\color{red}(d)} w/o RAFT \& TA Loss}      & \multicolumn{1}{c|}{9.61}      & \multicolumn{1}{c|}{0.434}   & 28.68 / 0.8274 / 2.003 \\
\multicolumn{1}{l|}{{\color{red}(e)} w/o TA Loss}              & \multicolumn{1}{c|}{9.61}      & \multicolumn{1}{c|}{0.434}  & 28.73 / 0.8288 / 1.956 \\ 
\multicolumn{1}{l|}{{\color{red}(f)} End-to-End Learning}      & \multicolumn{1}{c|}{9.61}      & \multicolumn{1}{c|}{0.434}   & 28.39 / 0.8190 / 2.152 \\ \hline
\multicolumn{4}{c}{Multi-Attention}                                                                                                                                             \\ \hline
\multicolumn{1}{l|}{{\color{red}(g)} self-attn \cite{zamir2022restormer} + SFT \cite{wang2018sftgan}}      & \multicolumn{1}{c|}{9.20}      & \multicolumn{1}{c|}{0.415}  & 28.50 / 0.8244 / 2.039 \\
\multicolumn{1}{l|}{{\color{red}(h)} CO attn + SFT \cite{wang2018sftgan}}        & \multicolumn{1}{c|}{9.20}      & \multicolumn{1}{c|}{0.416}  & 28.58 / 0.8262 / {\color{blue}1.938} \\
\multicolumn{1}{l|}{{\color{red}(i)} self-attn \cite{zamir2022restormer} + DA attn} & \multicolumn{1}{c|}{9.61}      & \multicolumn{1}{c|}{0.434}   & {\color{blue}28.80} / {\color{blue}0.8298} / 1.956 \\ \hline
\multicolumn{1}{l|}{{\color{red}(j)} Ours}                     & \multicolumn{1}{c|}{9.61}      & \multicolumn{1}{c|}{0.434}   & {\color{red}\textbf{28.83}} / {\color{red}\textbf{0.8315}} / {\color{red}\textbf{1.918}} \\
\toprule
\end{tabular}}
\end{center}
\vspace{-5mm}
\caption{Ablation study on the components in FMA-Net.}
\vspace{-5mm}
\label{tab:abla_arch}
\end{table}

\begin{figure}[h]
\centering
\includegraphics[width=8.3cm]{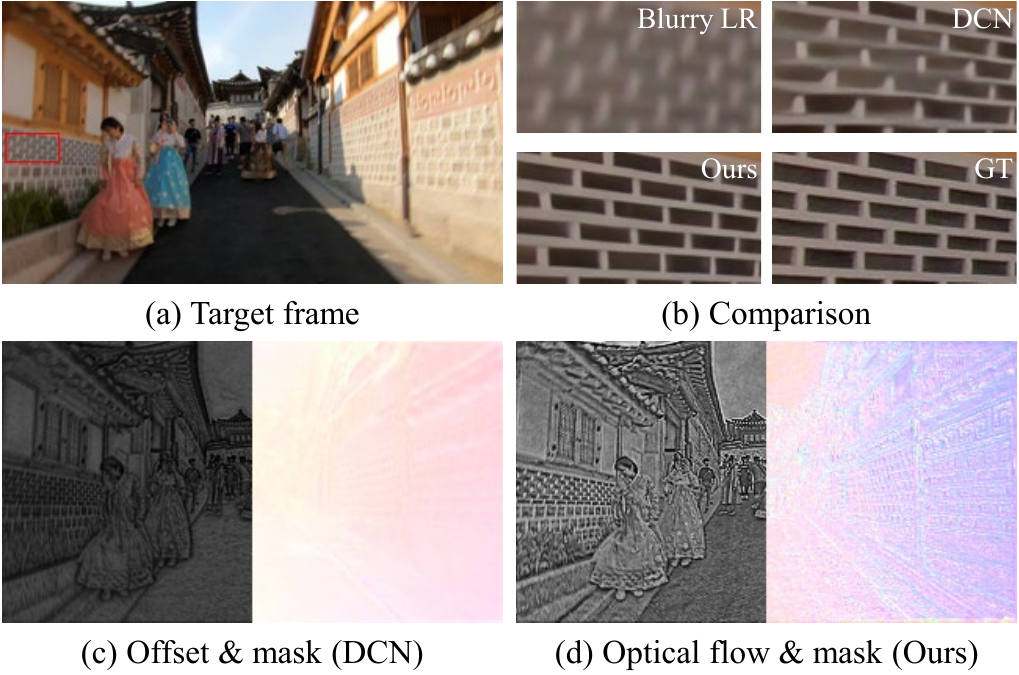}
\vspace{-4mm}
\caption{Offsets and mask (DCN \cite{dai2017deformable}) vs. Multi-flow-mask pairs $\textbf{f}$ (FMA-Net). Analysis of the multi-flow-mask pairs $\textbf{f}$ compared to the DCN \cite{dai2017deformable}. The offset and optical flow maps with their largest deviations are visualized with their corresponding masks.}
\vspace{-3mm}
\label{fig:dcn_comparison}
\end{figure}

\noindent \textbf{Design choices for FMA-Net.}\quad Table \ref{tab:abla_arch} shows the ablation experiment results for the components of our FMA-Net: 

\begin{figure}[t]
\centering
\includegraphics[width=8.3cm]{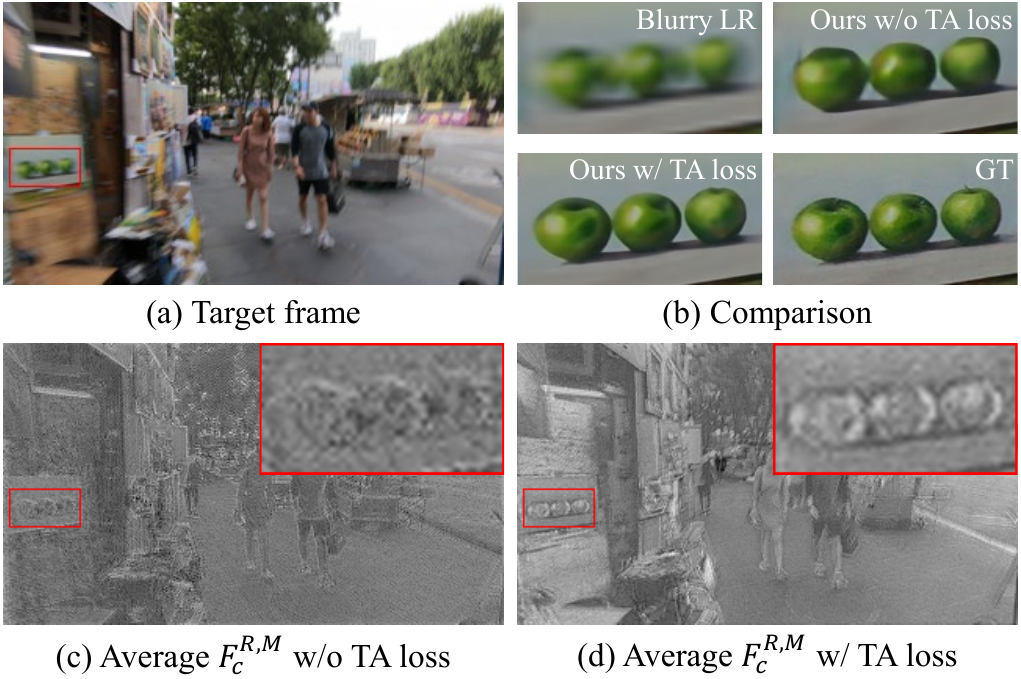}
\caption{Analysis of the TA loss.}
\vspace{-5mm}
\label{fig:ta_loss}
\end{figure}

\begin{itemize}
\vspace{0.5mm}
    \item [(i)] Table \ref{tab:abla_arch}{\color{red}(a-b, j)} shows the performance change in the number of multi-flow-mask pairs $n$. As $n$ increases, there is a significant performance improvement in $Net^R$, accompanied by a slight increase in memory cost. The best results are observed in Table \ref{tab:abla_arch}{\color{red}(j)} with $n=9$;
\vspace{0.5mm}
    \item [(ii)] Table \ref{tab:abla_arch}{\color{red}(c)} shows the result of implicitly utilizing motion information through DCN \cite{dai2017deformable} instead of using our multi-flow-mask pairs $\textbf{f}$. With the same number of offsets and $n$, our method achieves $0.31$ dB higher performance compared to using DCN. This is due to the utilization of the self-induced sharper optical flows and occlusion masks, as shown in Fig. \ref{fig:dcn_comparison};
\vspace{0.5mm}
    \item [(iii)] Table \ref{tab:abla_arch}{\color{red}(d-f, j)} shows the performance change depending on the used loss functions and training strategies. The `RAFT' in Table \ref{tab:abla_arch}{\color{red}(d)} refers to the use of $l_1(f^Y, f^Y_{RAFT})$ in Eq. \ref{eq:loss_d} for $Net^D$. The effectiveness of our loss functions in Eqs. \ref{eq:loss_d} and \ref{eq:loss_total} can be observed from Table \ref{tab:abla_arch}{\color{red}(d-e, j)}, especially with our new TA loss which anchors and sharpens each feature with respect to the corresponding frame index as shown in Fig. \ref{fig:ta_loss}, leading to $0.1$ dB PSNR improvement (Table \ref{tab:abla_arch}{\color{red}(e)} and {\color{red}(j)}). Also, our two-stage training strategy achieves $0.44$ dB improvement (Table \ref{tab:abla_arch}{\color{red}(f)});
    \item [(iv)] For the ablation study on our multi-attention (CO + DA attentions), we replaced them with self-attention \cite{zamir2022restormer} and spatial feature transform (SFT) \cite{wang2018sftgan} layer that is a SOTA feature modulation module, respectively. The results in Table \ref{tab:abla_arch}{\color{red}(g-j)} clearly demonstrate that our multi-attention approach outperforms the SOTA self-attention and modulation methods with a $0.33$ dB improvement.
\end{itemize}

\vspace{-0.5mm}
Similarly, $Net^D$ exhibits the same tendencies as $Net^R$. See the results and analysis in the \textit{Supplemental}.


\section{Conclusion}

We propose a novel VSRDB framework, called FMA-Net, based on our novel FGDF and FRMA. We iteratively update features including self-induced optical flow through stacked FRMA blocks, and predict a flow-mask pair with flow-guided dynamic filters, which enables the network to capture and utilize small-to-large motion information. The FGDF leads to a dramatic performance improvement compared to conventional dynamic filtering. Additionally, the newly proposed temporal anchor (TA) loss facilitates model training by temporally anchoring and sharpening unwarped features. Extensive experiments demonstrate that our FMA-Net achieves best performances for diverse datasets with significant margins compared to the recent SOTA methods.

\vspace{2mm}
\noindent \textbf{Acknowledgement.}\quad This work was supported by the IITP grant funded by the Korea government (MSIT): No. 2021-0-00087, Development of high-quality conversion technology for SD/HD low-quality media and No. RS2022-00144444, Deep Learning Based Visual Representational Learning and Rendering of Static and Dynamic Scenes.


\newpage
\FloatBarrier

{
    \small
    \bibliographystyle{ieeenat_fullname}
    \bibliography{main}
}

\clearpage
\setcounter{page}{1}
\maketitlesupplementary
\appendix

In this supplementary material, we first describe the ablation studies for various components of our design on FMA-Net in Sec. \ref{sec: supp_abla}. Subsequently, in Sec. \ref{sec: supp_exp}, we introduce a lightweight version of FMA-Net and present the performance of VSRDB methods and all possible combinations of sequential cascade approaches in REDS4 \cite{nah2019ntire}. Additionally, we also provide additional qualitative comparison results and a demo video. Finally, we discuss the limitations of our FMA-Net in Sec. \ref{sec: supp_disc}. 

We also recommend the readers to refer to our project page at \url{https://kaist-viclab.github.io/fmanet-site} where the source codes and the pre-trained models are available for the sake of \textit{reproducibility}.

\section{Ablation Studies} \label{sec: supp_abla}

\subsection{Effect of flow-guided dynamic filtering (FGDF)}
\begin{figure*}[t]
\centering
\includegraphics[width=15cm]{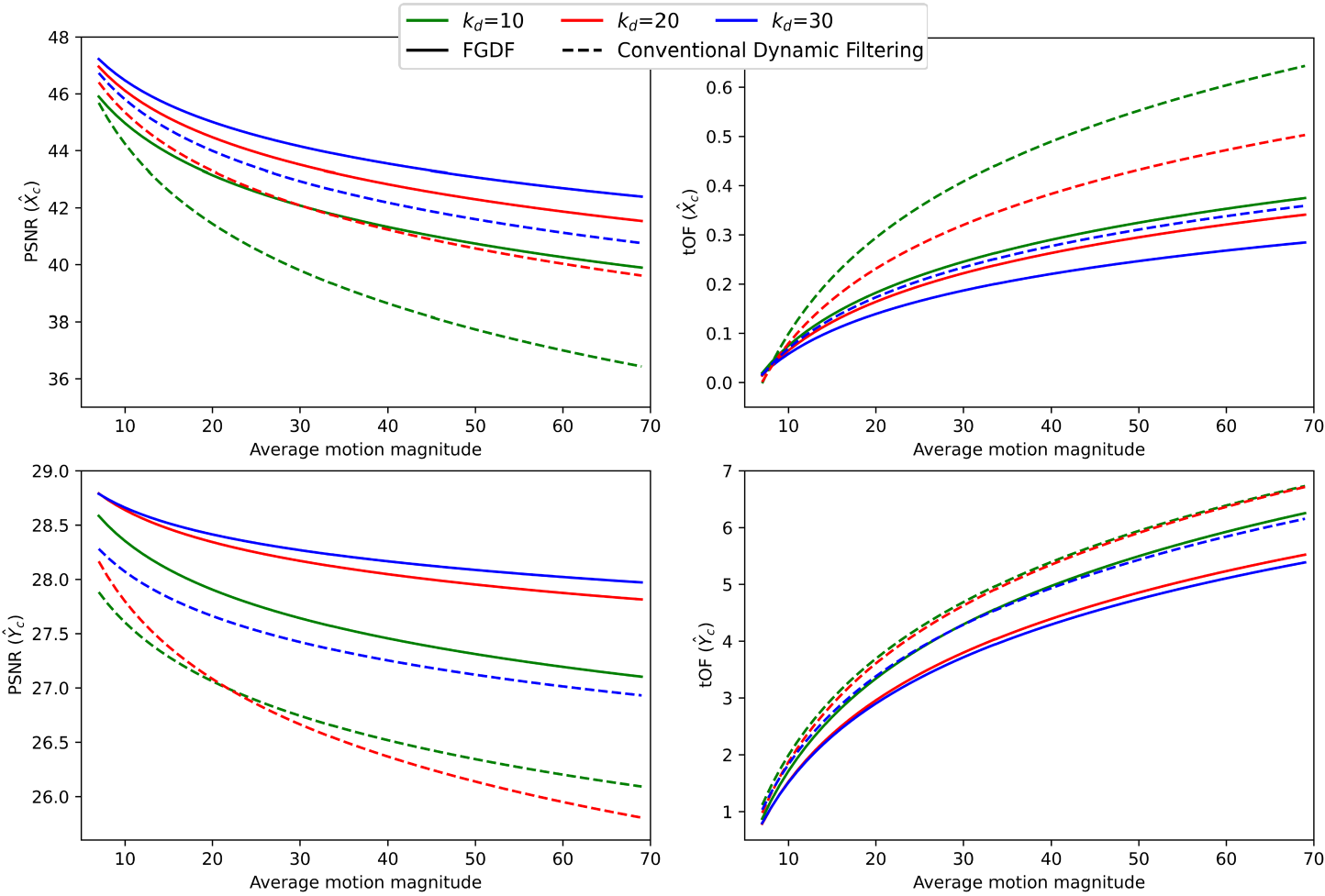}
\caption{Flow-guided dynamic filtering (FGDF) vs. conventional dynamic filtering \cite{jia2016dynamic, jo2018deep, kim2021koalanet}. Trendline visualization for Table \ref{tab:abla_fgdf} of the main paper.}
\label{fig:supp_trendline}
\end{figure*}

Fig. \ref{fig:supp_trendline} shows the spatial quality (PSNR) and temporal consistency (tOF) performance over average motion magnitudes which are also tabulated in Table \ref{tab:abla_fgdf} of the main paper. As shown in Fig. \ref{fig:supp_trendline}, our Flow-Guided Dynamic Filtering (FGDF) significantly outperforms the conventional dynamic filtering \cite{jia2016dynamic, jo2018deep, kim2021koalanet} in terms of PSNR and tOF metrics across all average motion magnitudes. It is noted that our FGDF gets more superior as the average motion magnitudes increase, indicating the effectiveness of FGDF, which is aware of motion trajectory, over the conventional dynamic filtering based on fixed positions and surroundings.

\subsection{Design choices for FMA-Net}
Table \ref{tab:abla_arch_supp} presents more detailed results of the ablation study from Table \ref{tab:abla_arch} of the main paper, additionally including the reconstruction performance of the degradation learning network $Net^D$. The tendency of performance changes on the selection of components for $Net^D$ are similar to those for $Net^R$, demonstrating the effectiveness of our multi-flow-mask pairs (Table \ref{tab:abla_arch_supp}{\color{red}(a-b, j)}, loss functions (Table \ref{tab:abla_arch_supp}{\color{red}(d-e, j)}), training strategy (Table \ref{tab:abla_arch_supp}{\color{red}(f, j)}), and multi-attention module (Table \ref{tab:abla_arch_supp}{\color{red}(g-j)}). It should be noted that the two reconstruction performances of $Net^D$ in Table \ref{tab:abla_arch_supp}{\color{red}(h)} (CO attn + SFT \cite{wang2018sftgan}) and Table \ref{tab:abla_arch_supp}{\color{red}(j)} (CO attn + DA attn) are the same because the SFT \cite{wang2018sftgan} and DA attn are only utilized in $Net^R$. The same tendency is also observed in Table \ref{tab:abla_arch_supp}{\color{red}(g,i)} because the same $NetD$ is used.

\subsection{Number of input frames}
Table \ref{tab:abla_num_frames} shows the performance of FMA-Net according to the different numbers of input frames $T$. It shows that as $T$ increases, the performance of both $Net^D$ and $Net^R$ improves, indicating that the FMA-Net effectively utilizes long-term information. Considering the trade-off between computational complexity and performance, we finally adopted $T=3$.

\subsection{Iterative Feature Refinement}
Fig. \ref{fig:supp_wf_refine} illustrates the iterative refinement process of the warped feature $F_w^{R,i}$ in FRMA blocks of $Net^R$ across three different scenes. In these scenes, it is evident that $F_w^{R,i}$ becomes sharper and more activated through iterative refinement, demonstrating the effectiveness of our iterative feature refinement with multi-attention (FRMA) block in improving the overall performance for VSRDB.

\subsection{Multi-flow-mask pairs}
\begin{figure*}[p]
\centering
\includegraphics[width=15cm]{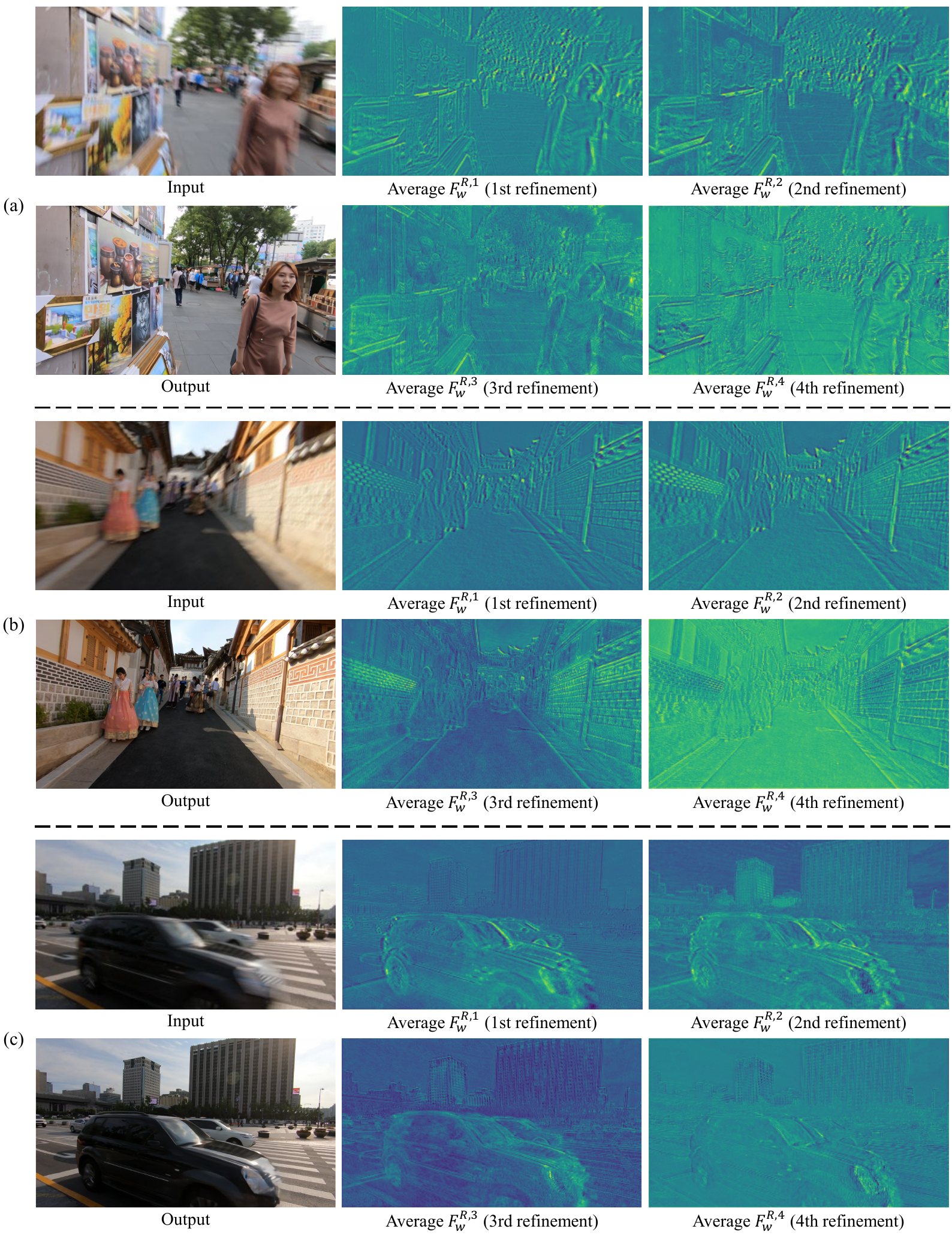}
\caption{Visualization of iterative refinement process of warped feature $F_w^{R,i}$ in FRMA blocks of $Net^R$. The brighter the pixel, the more activated it is.}
\vspace{-5mm}
\label{fig:supp_wf_refine}
\end{figure*}

\begin{figure*}[p]
\centering
\includegraphics[width=17cm]{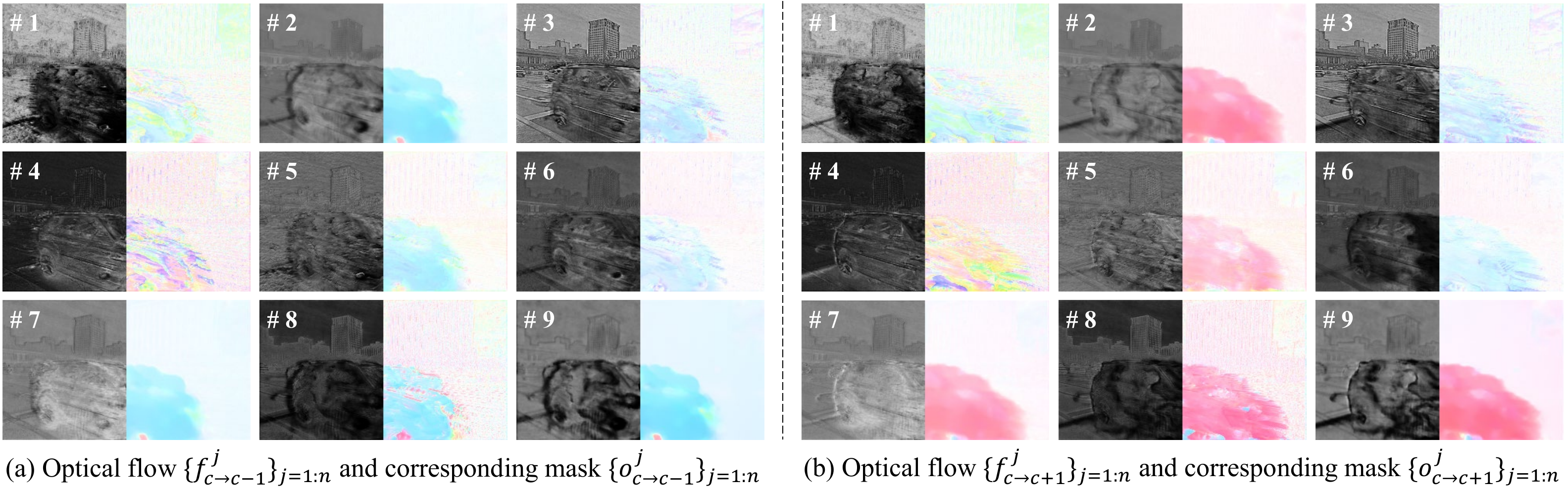}
\caption{Visualisation of multi-flow-mask pairs $\textbf{f}^{R,M}$ in $Net^R$.}
\vspace{-5mm}
\label{fig:supp_multi_flow}
\end{figure*}

Fig. \ref{fig:supp_multi_flow} illustrates an example of multi-flow-mask pairs $\textbf{f}^{R,M}$ in $Net^R$. In contrast to conventional sharp LR VSR methods \cite{chan2022basicvsr++, caballero2017real, tao2017detail, liu2022learning} that only utilize smooth optical flows with similar values among pixels belonging to the same object, the optical flows in Fig. \ref{fig:supp_multi_flow} include not only smooth optical flows (\# 2, \# 7, and \# 9 in Fig. \ref{fig:supp_multi_flow}) but also sharp optical flows (\# 1, \# 3-6, and \# 8 in Fig. \ref{fig:supp_multi_flow}) with varying values among pixels belonging to the same object. This distinction arises from our multi-flow-mask pairs $\textbf{f}$ not only \textit{align} features as in conventional VSR methods, but also \textit{sharpen} blurry features where the blur is pixel-wise-variant, even among pixels belonging to the same object. The smooth optical flows \textit{align} features, while the sharp optical flows \textit{sharpen} them. Fig. \ref{fig:supp_wf_refine}{\color{red}(c)} shows the iterative refinement process of the aligned and sharpened warped feature $F_w$ using the multi-flow-mask pairs $\textbf{f}$ in the same scene as Fig. \ref{fig:supp_multi_flow}, demonstrating the effectiveness of our multi-flow-mask pairs for VSRDB.

\subsection{Visualization of FGDF process}
\begin{figure*}
\centering
\includegraphics[width=15cm]{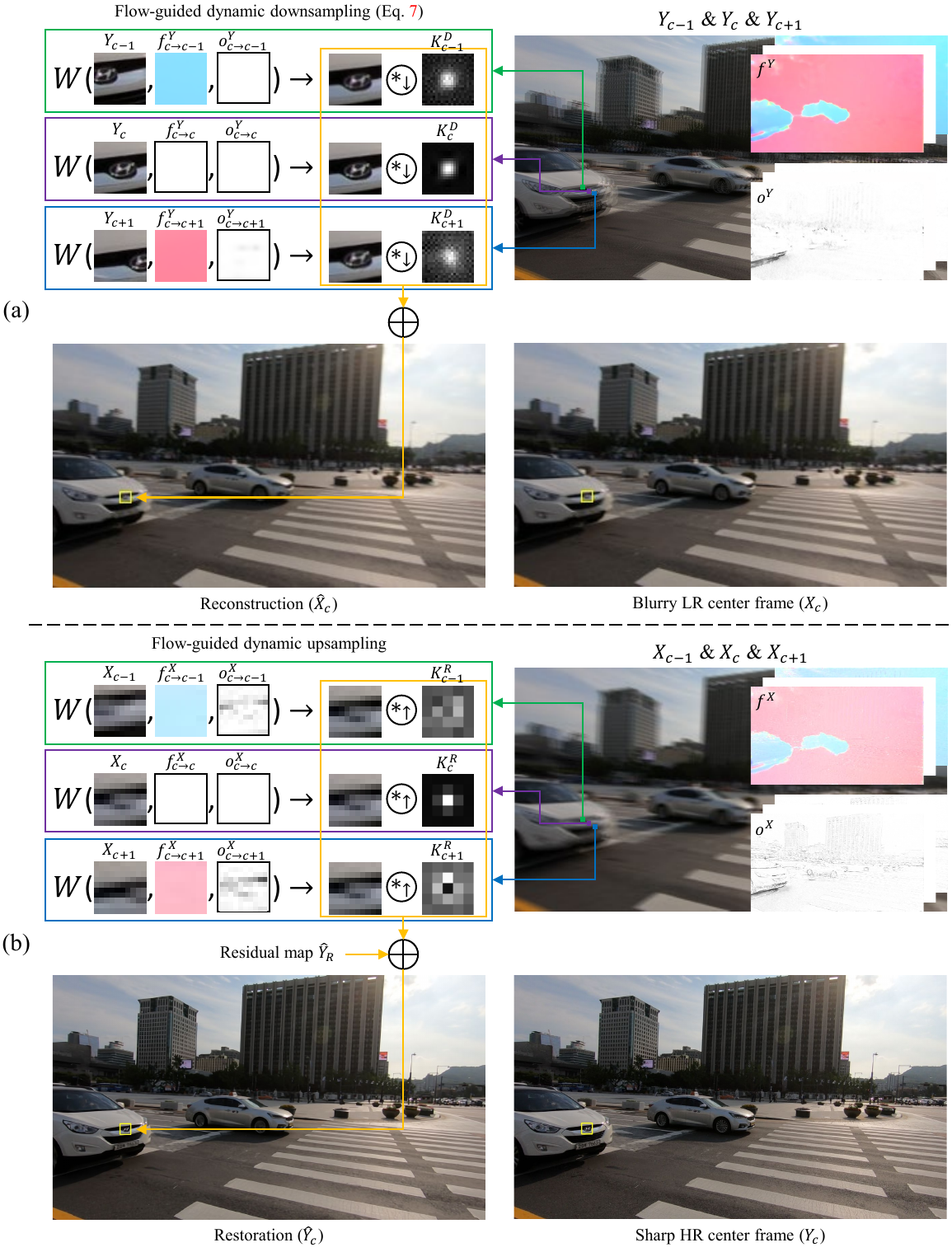}
\caption{Visualization of the flow-guided dynamic filtering (FGDF) process, including two image flow-mask pairs ($\textbf{f}^Y$ and $\textbf{f}^X$) and two dynamic kernels ($K^D$ and $K^R$): (a) Flow-guided dynamic downsampling (Eq. \ref{eq:recon} of the main paper) with spatio-temporally variant degradation kernels $K^D$; (b) Flow-guided dynamic upsampling with spatio-temporally variant restoration kernels $K^R$.}
\vspace{-5mm}
\label{fig:supp_fgdf_visualization}
\end{figure*}

Fig. \ref{fig:supp_fgdf_visualization} illustrates the proposed flow-guided dynamic filtering (FGDF) process, where Fig. \ref{fig:supp_fgdf_visualization}{\color{red}(a)} shows the flow-guided dynamic downsampling process of $Net^D$, and Fig. \ref{fig:supp_fgdf_visualization}{\color{red}(b)} illustrates the flow-guided dynamic upsampling process of $Net^R$. In particular, in Fig. \ref{fig:supp_fgdf_visualization}{\color{red}(a)}, the two degradation kernels of the neighboring frames tend to have peaky values around their own centers, because $Net^D$ filters a sharp HR sequence $Y_w$ aligned to the center frame index $c$ based on the image-mask pair $\textbf{f}^Y$ for $Y$. This allows $Net^D$ to effectively handle large motions with relatively small-sized kernels, as demonstrated in Fig. \ref{fig:supp_trendline} and Table \ref{tab:abla_fgdf} of the main paper. Similar to $Net^D$, $Net^R$ filters the aligned blurry LR sequence $X_w$ to the center frame index $c$ by the image flow-mask pair $\textbf{f}^X$ for $X$. We normalized the restoration kernels $K^R$ such that their kernel weights are allowed to take on positive and negative values, where the negative kernel weights can facilitate the deblurring process (dark regions of the kernels in Fig. \ref{fig:supp_fgdf_visualization}{\color{red}(b)} represent negative values), similar to \cite{kim2021koalanet}. We empirically found that this approach can restore the low-frequencies more effectively than simple interpolation methods such as bilinear and bicubic interpolations. Combining these restored low frequencies with the high-frequency details $\hat{Y}_r$ predicted by $Net^R$ in a residual learning manner results in faster training convergence and better performance compared to residual learning with bicubic upsampling or without residual learning.

\subsection{The Number of FRMA Blocks $M$}
Table \ref{tab:abla_num_M} shows the performance of FMA-Net according to the different numbers of FRMA Blocks $M$. It shows that as $M$ increases, the performance of FMA-Net improves, indicating that the stacked FRMA blocks can effectively update features. Besides Table \ref{tab:abla_num_M}, as can also be seen in Table \ref{tab:reds4_comparison} of the main paper, our \textit{smallest} FMA-Net variant ($M=1$) even shows superior performance than the previous SOTA methods.

\section{Detailed Experimental Results} \label{sec: supp_exp}
\subsection{FMA-Net$_s$}
We first introduce FMA-Net$_s$, a lightweight model of FMA-Net. FMA-Net$_s$ is a model that changes the number of FRMA blocks, $M$, from the original $4$ to $2$, with no other modifications. Table \ref{tab:supp_reds4} compares the quantitative performance of FMA-Net$_s$ on REDS4 \cite{nah2019ntire} dataset with one VSRDB method (HOFFR \cite{fang2022high}), four retrained SOTA methods (Restormer$^*$ \cite{zamir2022restormer}, GShiftNet$^*$ \cite{li2023simple}, BasicVSR++$^*$ \cite{chan2022basicvsr++}
, and RVRT$^*$ \cite{liang2022recurrent}
) for VSRDB on REDS \cite{nah2019ntire}, and our FMA-Net. Our FMA-Net$_s$ demonstrates the second-best performance, maintaining performance while reducing memory usage and runtime.

\subsection{Clip-by-clip Results on REDS4}
Table \ref{tab:supp_reds4_all_exp} shows the performance of the clip-by-clip results on REDS4 \cite{nah2019ntire} for VSRDB methods and all possible combinations of the sequential cascade approaches. It shows that our FMA-Net exhibits the best performance on all REDS4 clips consisting of realistic and dynamic scenes. In particular, compared to RVRT$^*$ \cite{liang2022recurrent}, our FMA-Net achieves PSNR improvement of 0.35 dB in Clip 000, a scene with small motion, improvements of 1.62 dB and 1.58 dB in Clips 011 and 020, scenes with large motion, respectively. This demonstrates the superiority of FMA-Net over existing SOTA methods, especially in scenes with large motion.

\subsection{Visualization Results}
\begin{figure*}[p]
\centering
\includegraphics[width=15cm]{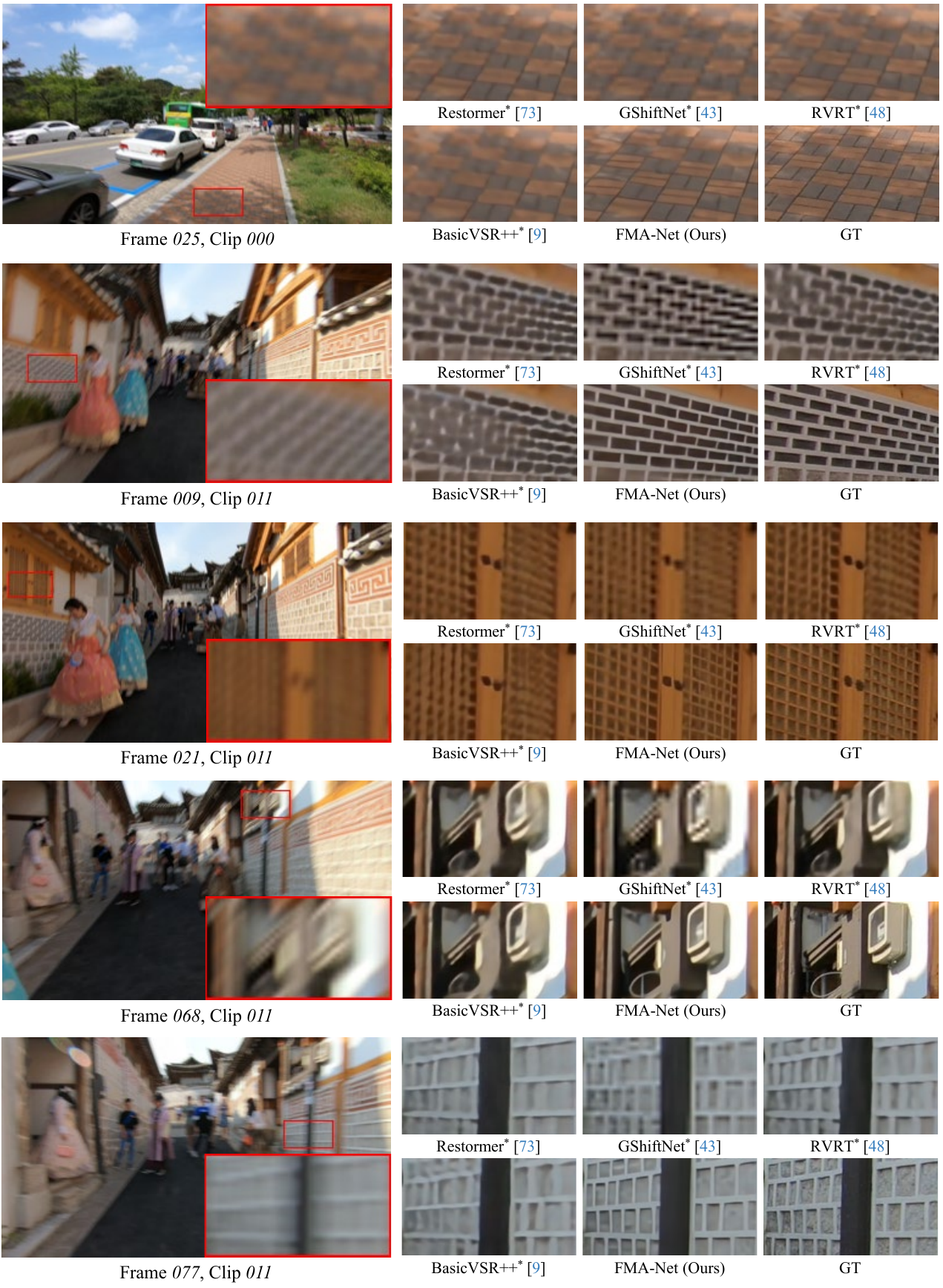}
\caption{Visual results of different methods on REDS4 \cite{nah2019ntire}. \textit{Best viewed in zoom}.}
\vspace{-5mm}
\label{fig:supp_reds4}
\end{figure*}

\begin{figure*}[p]
\centering
\includegraphics[width=15cm]{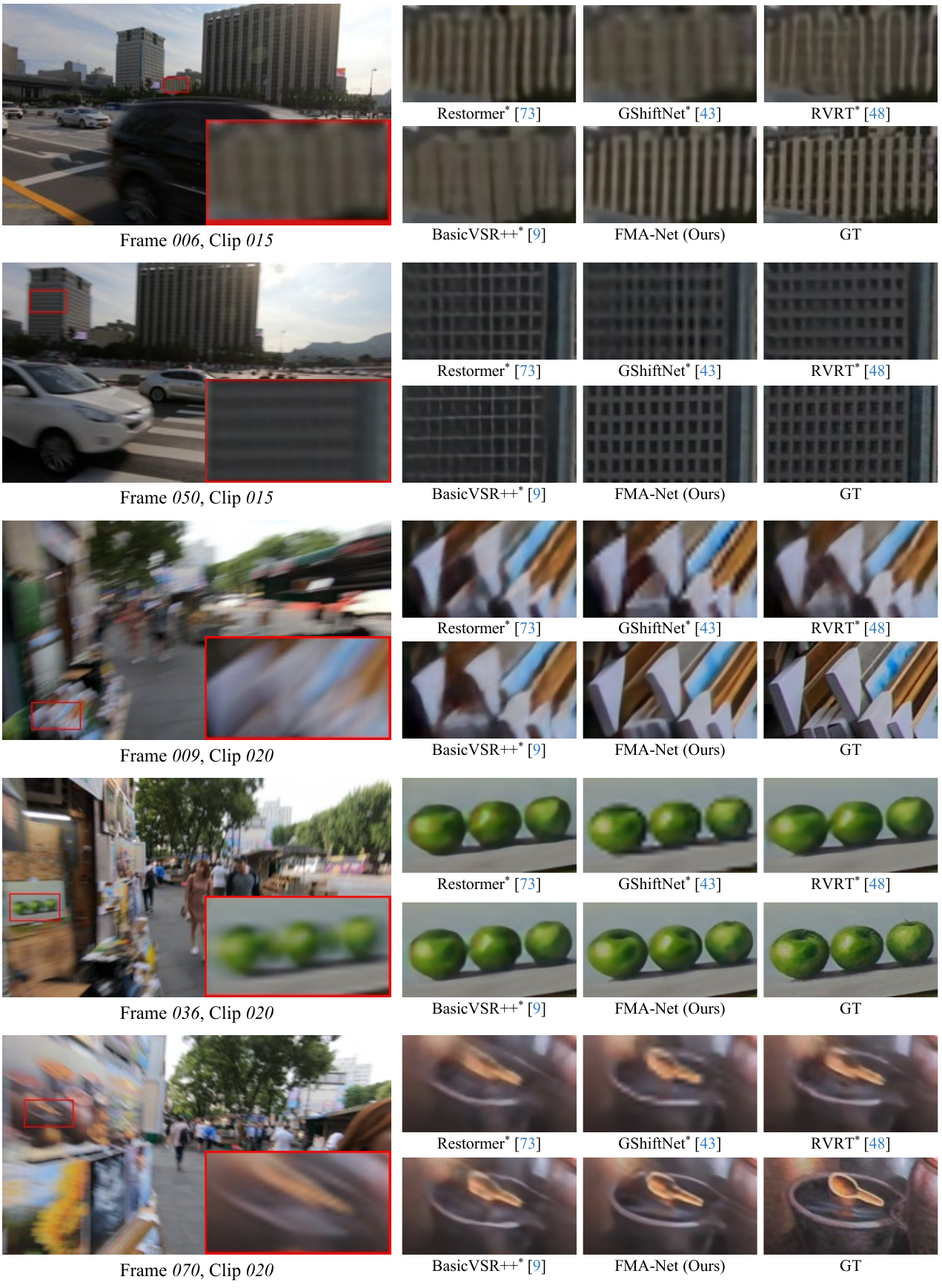}
\caption{Visual results of different methods on REDS4 \cite{nah2019ntire}. \textit{Best viewed in zoom}.}
\vspace{-5mm}
\label{fig:supp_reds4_2}
\end{figure*}

\begin{figure*}[p]
\centering
\includegraphics[width=15cm]{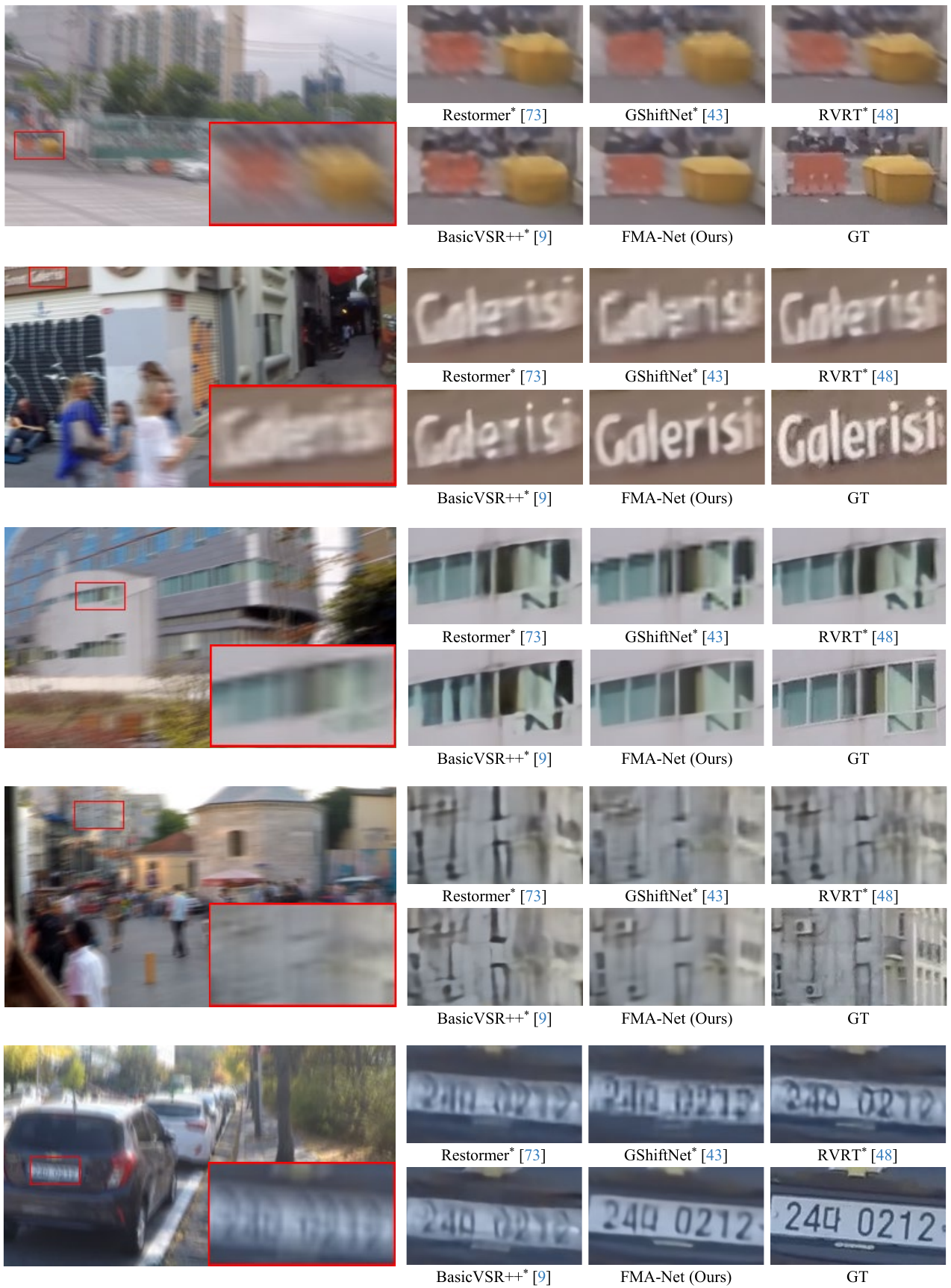}
\caption{Visual results of different methods on GoPro \cite{nah2019ntire} test set. \textit{Best viewed in zoom}.}
\vspace{-5mm}
\label{fig:supp_gopro}
\end{figure*}

We show more qualitative comparison results among the proposed FMA-Net and other SOTA methods on two benchmark datasets. The results for REDS4 \cite{nah2019ntire} and GoPro \cite{nah2017deep} are shown in Figs. \ref{fig:supp_reds4}-\ref{fig:supp_reds4_2} and Fig. \ref{fig:supp_gopro}, respectively.

\subsection{Visual Comparisons with Demo Video}
We provide a video at \url{https://www.youtube.com/watch?v=kO7KavOH6vw} to compare our FMA-Net with existing SOTA methods \cite{liang2022recurrent, chan2022basicvsr++, li2023simple}. The demo video includes comparisons between FMA-Net and SOTA methods on two clips from the REDS4 \cite{nah2019ntire} dataset and one clip from the GoPro \cite{nah2017deep} dataset.

\section{Discussions} \label{sec: supp_disc}

\subsection{Learning Scheme}
We train FMA-Net in a 2-stage manner which requires additional training time rather than end-to-end. This choice is made because, during the multi-attention process of $Net^R$, the warped feature $F_w$ is adjusted by the predicted degradation from $Net^D$ in a globally adaptive manner. When the network is trained end-to-end, in the initial training stages, $F_w$ is adjusted for incorrectly predicted kernels due to the random initialization of weights, which adversely affects the training process (The performance comparison between end-to-end and 2-stage strategies can be found in Table \ref{tab:abla_arch_supp}{\color{red}(f, j)})). To address this, we adopt a pre-training strategy for $Net^D$, which inevitably leads to longer training times compared to the end-to-end approach.

\subsection{Limitation: Object Rotation}
\begin{figure*}[p]
\centering
\includegraphics[width=15cm]{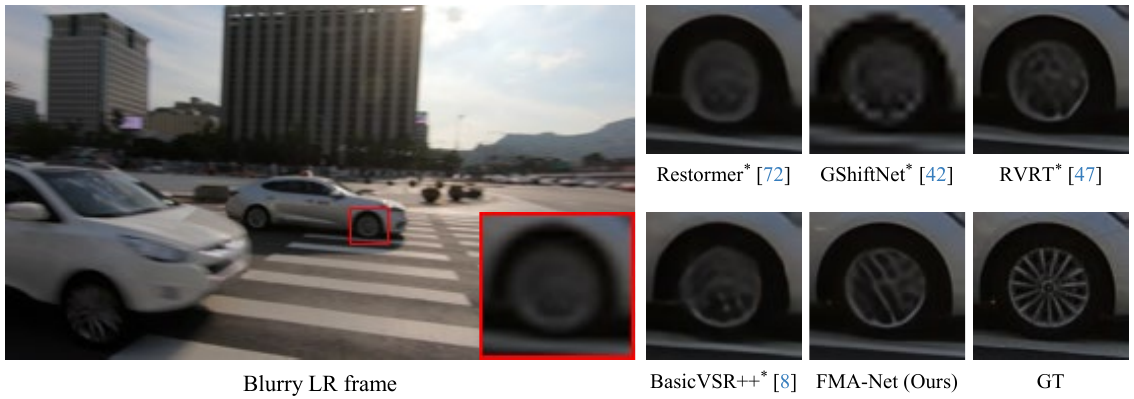}
\caption{Qualitative comparison for the extreme scene including object rotation.}
\vspace{-5mm}
\label{fig:supp_limit}
\end{figure*}

In extreme conditions such as object rotation, it is challenging to predict accurate optical flow, making precise restoration difficult. Fig. \ref{fig:supp_limit} illustrates the restoration results in a scene with object rotation, showing the failure of all methods, including our FMA-Net, in restoring a rotating object. The introduction of learnable homography parameters or the adoption of quaternion representations could be one option to enhance the performance in handling rotational motions.

\renewcommand{\arraystretch}{1.1}
\begin{table*}[h]
\begin{center}
\scalebox{0.9}{
\begin{tabular}{lcccc}
\bottomrule
\multicolumn{1}{c|}{\multirow{2}{*}{Methods}}  & \multicolumn{1}{c|}{\# Params} & \multicolumn{1}{c|}{Runtime} & \multicolumn{1}{c|}{$Net^D$ (blurry LR $\hat{X}_c$)}                   & $Net^R$ (sharp HR $\hat{Y}_c$)                  \\
\multicolumn{1}{c|}{}                         & \multicolumn{1}{c|}{(M)}       & \multicolumn{1}{c|}{(s)}     & \multicolumn{1}{c|}{PSNR $\uparrow$ / SSIM $\uparrow$ / tOF $\downarrow$}      & PSNR $\uparrow$ / SSIM $\uparrow$ / tOF $\downarrow$      \\ \hline
\multicolumn{5}{c}{The number of multi-flow-mask pairs $n$}                    \\ \hline
\multicolumn{1}{l|}{{\color{red}(a)} $n=1$}                      & \multicolumn{1}{c|}{9.15}      & \multicolumn{1}{c|}{0.424}   & \multicolumn{1}{c|}{44.80 / 0.9955 / 0.096} & 28.24 / 0.8151 / 2.224 \\
\multicolumn{1}{l|}{{\color{red}(b)} $n=5$}                      & \multicolumn{1}{c|}{9.29}      & \multicolumn{1}{c|}{0.429}   & \multicolumn{1}{c|}{{\color{blue}45.37} / {\color{blue}0.9960} / 0.086} & 28.60 / 0.8258 / 2.054 \\ \hline
\multicolumn{5}{c}{Deformable Convolutions \cite{dai2017deformable}}                                                                                                                                           \\ \hline
\multicolumn{1}{l|}{{\color{red}(c)} w/ DCNs (\#$\mbox{offset}=9$)}         & \multicolumn{1}{c|}{10.13}     & \multicolumn{1}{c|}{0.426}   & \multicolumn{1}{c|}{45.17 / 0.9956 / 0.093} & 28.52 / 0.8225 / 2.058 \\ \hline
\multicolumn{5}{c}{Loss Function and Training Strategy}                                                                                                                                                    \\ \hline
\multicolumn{1}{l|}{{\color{red}(d)} w/o RAFT \& TA Loss}      & \multicolumn{1}{c|}{9.62}      & \multicolumn{1}{c|}{0.434}   & \multicolumn{1}{c|}{45.28 / 0.9958 / 0.084} & 28.68 / 0.8274 / 2.003 \\
\multicolumn{1}{l|}{{\color{red}(e)} w/o TA Loss}              & \multicolumn{1}{c|}{9.62}      & \multicolumn{1}{c|}{0.434}   & \multicolumn{1}{c|}{45.33 / 0.9959 / {\color{blue}0.083}} & 28.73 / 0.8288 / 1.956 \\ 
\multicolumn{1}{l|}{{\color{red}(f)} End-to-End Learning}      & \multicolumn{1}{c|}{9.62}      & \multicolumn{1}{c|}{0.434}   & \multicolumn{1}{c|}{44.14 / 0.9947 / 0.107}              & 28.39 / 0.8190 / 2.152 \\ \hline
\multicolumn{5}{c}{Multi-Attention}                                                                                                                                             \\ \hline
\multicolumn{1}{l|}{{\color{red}(g)} self-attn \cite{zamir2022restormer} + SFT \cite{wang2018sftgan}}      & \multicolumn{1}{c|}{9.20}      & \multicolumn{1}{c|}{0.415}   & \multicolumn{1}{c|}{{\color{blue}45.37} / 0.9959 / 0.085} & 28.50 / 0.8244 / 2.039 \\
\multicolumn{1}{l|}{{\color{red}(h)} CO attn + SFT \cite{wang2018sftgan}}        & \multicolumn{1}{c|}{9.20}      & \multicolumn{1}{c|}{0.416}   & \multicolumn{1}{c|}{{\color{red}\textbf{45.46}} / {\color{red}\textbf{0.9961}} / {\color{red}\textbf{0.082}}} & 28.58 / 0.8262 / {\color{blue}1.938} \\
\multicolumn{1}{l|}{{\color{red}(i)} self-attn \cite{zamir2022restormer} + DA attn} & \multicolumn{1}{c|}{9.62}      & \multicolumn{1}{c|}{0.434}   & \multicolumn{1}{c|}{{\color{blue}45.37} / 0.9959 / 0.085} & {\color{blue}28.80} / {\color{blue}0.8298} / 1.956 \\ \hline
\multicolumn{1}{l|}{{\color{red}(j)} Ours}                     & \multicolumn{1}{c|}{9.62}      & \multicolumn{1}{c|}{0.434}   & \multicolumn{1}{c|}{{\color{red}\textbf{45.46}} / {\color{red}\textbf{0.9961}} / {\color{red}\textbf{0.082}}} & {\color{red}\textbf{28.83}} / {\color{red}\textbf{0.8315}} / {\color{red}\textbf{1.918}} \\
\toprule
\end{tabular}}
\end{center}
\vspace{-0.5cm}
\caption{Ablation study on the components in FMA-Net.}
\label{tab:abla_arch_supp}
\end{table*}

\renewcommand{\arraystretch}{1.1}
\begin{table*}
\begin{center}
\setlength\tabcolsep{15pt} 
\scalebox{0.9}{
\begin{tabular}{c|c|c|c|c}
\bottomrule
\multirow{2}{*}{$T$} & \# Params & Runtime & $Net^D$    & $Net^R$    \\
                   & (M)                              &  (s)                            & PSNR $\uparrow$ / SSIM $\uparrow$ / tOF $\downarrow$      & PSNR $\uparrow$ / SSIM $\uparrow$ / tOF $\downarrow$     \\ \hline
1                  & 9.03                           & 0.206                        & 42.04 / 0.9908 / 0.182 & 27.33 / 0.7866 / 2.672 \\
3                  & 9.62                           & 0.434                        & 45.46 / 0.9961 / 0.082 & 28.83 / 0.8315 / 1.918 \\
5                  & 9.94                           & 0.737                        & {\color{blue}45.74} / {\color{blue}0.9965} / {\color{blue}0.076} & {\color{blue}28.92} / {\color{blue}0.8347} / {\color{blue}1.909} \\
7                  & 16.61                          & 1.425                        & {\color{red}\textbf{46.24}} / {\color{red}\textbf{0.9969}} / {\color{red}\textbf{0.068}} & {\color{red}\textbf{29.00}} / {\color{red}\textbf{0.8376}} / {\color{red}\textbf{1.856}} \\
\toprule
\end{tabular}}
\end{center}
\vspace{-5mm}
\caption{Ablation study on the number of input frames $T$.}
\label{tab:abla_num_frames}
\end{table*}

\begin{table*}[h]
\begin{center}
\setlength\tabcolsep{15pt} 
\scalebox{0.9}{
\begin{tabular}{c|c|c|c}
\bottomrule
$M$ & \# Params (M) & Runtime (s) & PSNR $\uparrow$ / SSIM $\uparrow$ / tOF $\downarrow$     \\ \hline
1 & 6.3           & 0.147       & 28.07 / 0.8109 / 2.24 \\
2 & 7.4           & 0.231       & {\color{blue}28.46} / {\color{blue}0.8212} / {\color{blue}2.08} \\
4 & 9.6           & 0.427       & {\color{red}\textbf{28.83}} / {\color{red}\textbf{0.8315}} / {\color{red}\textbf{1.92}} \\ \toprule
\end{tabular}}
\end{center}
\vspace{-5mm}
\caption{Ablation study on the number of FRMA blocks $M$.}
\label{tab:abla_num_M}
\end{table*}

\renewcommand{\arraystretch}{1.0}
\begin{table*}[]
\begin{center}
\setlength\tabcolsep{20pt} 
\scalebox{0.9}{
\begin{tabular}{cccc}
\bottomrule
\multicolumn{1}{c|}{\multirow{2}{*}{Methods}} & \multicolumn{1}{c|}{\multirow{2}{*}{\# Params (M)}} & \multicolumn{1}{c|}{\multirow{2}{*}{Runtime (s)}} & REDS4             \\
\multicolumn{1}{c|}{}                         & \multicolumn{1}{c|}{}                               & \multicolumn{1}{c|}{}                             & PSNR $\uparrow$ / SSIM $\uparrow$ / tOF $\downarrow$ \\ 
\hline
\multicolumn{1}{c|}{HOFFR \cite{fang2022high}}                    & \multicolumn{1}{c|}{3.5}                              & \multicolumn{1}{c|}{0.500}                            & 27.24 / 0.7870 / -                 \\
\multicolumn{1}{c|}{Restormer$^*$ \cite{zamir2022restormer}}               & \multicolumn{1}{c|}{26.5}                              & \multicolumn{1}{c|}{0.081}                            & 27.29 / 0.7850 / 2.71                 \\
\multicolumn{1}{c|}{GShiftNet$^*$ \cite{li2023simple}}               & \multicolumn{1}{c|}{13.5}                              & \multicolumn{1}{c|}{0.185}                            & 25.77 / 0.7275 / 2.96                 \\ 
\multicolumn{1}{c|}{BasicVSR++$^*$ \cite{chan2022basicvsr++}}               & \multicolumn{1}{c|}{7.3}                              & \multicolumn{1}{c|}{0.072}                            & 27.06 / 0.7752 / 2.70                 \\
\multicolumn{1}{c|}{RVRT$^*$ \cite{liang2022recurrent}}                    & \multicolumn{1}{c|}{12.9}                              & \multicolumn{1}{c|}{0.680}                            & 27.80 / 0.8025 / 2.40                 \\
\hline

\multicolumn{1}{c|}{FMA-Net$_s$ (Ours)}                     & \multicolumn{1}{c|}{7.4}                              & \multicolumn{1}{c|}{0.231}                            & {\color{blue}28.46} / {\color{blue}0.8212} / {\color{blue}2.08}          \\ 

\multicolumn{1}{c|}{FMA-Net (Ours)}                     & \multicolumn{1}{c|}{9.6}                              & \multicolumn{1}{c|}{0.427}                            & {\color{red}\textbf{28.83}} / {\color{red}\textbf{0.8315}} / {\color{red}\textbf{1.92}}          \\ 

\toprule
\end{tabular}}
\end{center}
\vspace{-5mm}
\caption{Quantitative comparison on REDS4 for $\times 4$ VSRDB. All results are calculated on the RGB channel. {\color{red}\textbf{Red}} and {\color{blue}blue} colors indicate the best and second-best performance, respectively. Runtime is calculated on an LR frame sequence of size $180 \times 320$. The superscript $^*$ indicates that the model is retrained on the REDS \cite{nah2019ntire} training dataset for VSRDB.}
\vspace{-5mm}
\label{tab:supp_reds4}
\end{table*}

\renewcommand{\arraystretch}{1.1}
\begin{table*}[t]
\begin{center}
\scalebox{0.68}{
\begin{tabular}{ccllcllcllcllcll}
\bottomrule
\multicolumn{1}{c|}{REDS4}                              & \multicolumn{3}{c|}{CLIP 000}                & \multicolumn{3}{c|}{CLIP 011}                & \multicolumn{3}{c|}{CLIP 015}                & \multicolumn{3}{c|}{CLIP 020}                & \multicolumn{3}{c}{Average}                 \\ \cline{1-1}
\multicolumn{1}{c|}{Methods}                            & \multicolumn{3}{c|}{PSNR $\uparrow$ / SSIM $\uparrow$ / tOF $\downarrow$}       & \multicolumn{3}{c|}{PSNR $\uparrow$ / SSIM $\uparrow$ / tOF $\downarrow$}       & \multicolumn{3}{c|}{PSNR $\uparrow$ / SSIM $\uparrow$ / tOF $\downarrow$}       & \multicolumn{3}{c|}{PSNR $\uparrow$ / SSIM $\uparrow$ / tOF $\downarrow$}       & \multicolumn{3}{c}{PSNR $\uparrow$ / SSIM $\uparrow$ / tOF $\downarrow$}       \\ \hline
\multicolumn{16}{c}{Single Image Super-Resolution + Deblurring}                                                                                                                                                                                                                                       \\ \hline
\multicolumn{1}{c|}{Bicubic + Restormer \cite{zamir2022restormer}}                & \multicolumn{3}{c|}{24.16 / 0.6488 /   0.95} & \multicolumn{3}{c|}{22.92 / 0.6341 /   7.17} & \multicolumn{3}{c|}{26.14 / 0.7565 /   4.73} & \multicolumn{3}{c|}{21.24 / 0.6086 /   8.12} & \multicolumn{3}{c}{23.62 / 0.6620 /   5.24} \\
\multicolumn{1}{c|}{Bicubic + FFTformer \cite{kong2023efficient}}                & \multicolumn{3}{c|}{24.17 / 0.6432 / 0.90}   & \multicolumn{3}{c|}{22.90 / 0.6328 / 7.06}   & \multicolumn{3}{c|}{26.55 / 0.7669 / 4.85}   & \multicolumn{3}{c|}{21.27 / 0.6082 / 8.09}   & \multicolumn{3}{c}{23.72 / 0.6628 / 5.23}   \\
\multicolumn{1}{c|}{Bicubic + RVRT \cite{liang2022recurrent}}                     & \multicolumn{3}{c|}{24.21 / 0.6486 / 0.84}   & \multicolumn{3}{c|}{23.53 / 0.6818 / 4.87}   & \multicolumn{3}{c|}{26.58 / 0.7725 / 4.28}   & \multicolumn{3}{c|}{21.96 / 0.6641 / 5.90}   & \multicolumn{3}{c}{24.07 / 0.6918 / 3.97}   \\
\multicolumn{1}{c|}{Bicubic + GShiftNet \cite{li2023simple}}                & \multicolumn{3}{c|}{24.19 / 0.6468 / 0.80}   & \multicolumn{3}{c|}{23.36 / 0.6659 / 5.36}   & \multicolumn{3}{c|}{26.59 / 0.7742 / 4.31}   & \multicolumn{3}{c|}{21.76 / 0.6451 / 6.25}   & \multicolumn{3}{c}{23.98 / 0.6830 / 4.18}   \\ \hline
\multicolumn{1}{c|}{SwinIR \cite{liang2021swinir} + Restormer \cite{zamir2022restormer}}                 & \multicolumn{3}{c|}{25.22 / 0.7136 / 0.68}   & \multicolumn{3}{c|}{23.17 / 0.6566 / 6.65}   & \multicolumn{3}{c|}{27.49 / 0.8142 / 4.18}   & \multicolumn{3}{c|}{21.47 / 0.6316 / 7.78}   & \multicolumn{3}{c}{24.33 / 0.7040 / 4.82}   \\
\multicolumn{1}{c|}{SwinIR \cite{liang2021swinir} \cite{liang2021swinir} + FFTformer \cite{kong2023efficient}}                 & \multicolumn{3}{c|}{25.04 / 0.7096 / 0.66}   & \multicolumn{3}{c|}{23.06 / 0.6629 / 5.86}   & \multicolumn{3}{c|}{27.22 / 0.8183 / 3.94}   & \multicolumn{3}{c|}{21.40 / 0.6329 / 7.40}   & \multicolumn{3}{c}{24.18 / 0.7059 / 4.47}   \\
\multicolumn{1}{c|}{SwinIR \cite{liang2021swinir} + RVRT \cite{liang2022recurrent}}                      & \multicolumn{3}{c|}{25.32 / 0.7261 / 0.58}   & \multicolumn{3}{c|}{23.97 / 0.7317 / 4.00}   & \multicolumn{3}{c|}{27.36 / 0.8232 / 3.64}   & \multicolumn{3}{c|}{22.11 / 0.6995 / 5.61}   & \multicolumn{3}{c}{24.69 / 0.7451 / 3.46}   \\
\multicolumn{1}{c|}{SwinIR \cite{liang2021swinir} + GShiftNet \cite{li2023simple}}                 & \multicolumn{3}{c|}{25.30 / 0.7221 / 0.58}   & \multicolumn{3}{c|}{23.59 / 0.6964 / 4.78}   & \multicolumn{3}{c|}{27.38 / 0.8219 / 3.67}   & \multicolumn{3}{c|}{21.81 / 0.6687 / 5.94}   & \multicolumn{3}{c}{24.52 / 0.7273 / 3.74}   \\ \hline
\multicolumn{1}{c|}{HAT \cite{chen2023activating} + Restormer \cite{zamir2022restormer}}                    & \multicolumn{3}{c|}{25.21 / 0.7151 / 0.68}   & \multicolumn{3}{c|}{23.18 / 0.6579 / 6.55}   & \multicolumn{3}{c|}{27.57 / 0.8184 / 4.07}   & \multicolumn{3}{c|}{21.48 / 0.6323 / 7.74}   & \multicolumn{3}{c}{24.36 / 0.7059 / 4.76}   \\
\multicolumn{1}{c|}{HAT \cite{chen2023activating} + FFTformer \cite{kong2023efficient}}                    & \multicolumn{3}{c|}{25.11 / 0.7136 / 0.66}   & \multicolumn{3}{c|}{23.12 / 0.6670 / 5.75}   & \multicolumn{3}{c|}{27.26 / 0.8208 / 3.85}   & \multicolumn{3}{c|}{21.42 / 0.6348 / 7.35}   & \multicolumn{3}{c}{24.22 / 0.7091 / 4.40}   \\
\multicolumn{1}{c|}{HAT \cite{chen2023activating} + RVRT \cite{liang2022recurrent}}                         & \multicolumn{3}{c|}{25.39 / 0.7299 / 0.57}   & \multicolumn{3}{c|}{24.03 / 0.7354 / 3.93}   & \multicolumn{3}{c|}{27.48 / 0.8289 / 3.53}   & \multicolumn{3}{c|}{22.15 / 0.7022 / 5.54}   & \multicolumn{3}{c}{24.76 / 0.7491 / 3.39}   \\
\multicolumn{1}{c|}{HAT \cite{chen2023activating} + GShiftNet \cite{li2023simple}}                    & \multicolumn{3}{c|}{25.36 / 0.7256 / 0.57}   & \multicolumn{3}{c|}{23.63 / 0.6989 / 4.73}   & \multicolumn{3}{c|}{27.48 / 0.8270 / 3.60}   & \multicolumn{3}{c|}{21.83 / 0.6699 / 5.91}   & \multicolumn{3}{c}{24.58 / 0.7304 / 3.70}   \\ \hline
\multicolumn{16}{c}{Video Super-Resolution + Deblurring}                                                                                                                                                                                                                                              \\ \hline
\multicolumn{1}{c|}{BasicVSR++ \cite{chan2022basicvsr++} + Restormer \cite{zamir2022restormer}}             & \multicolumn{3}{c|}{26.35 / 0.7765 /   0.48} & \multicolumn{3}{c|}{23.08 / 0.6527 /   6.83} & \multicolumn{3}{c|}{28.07 / 0.8421 /   3.78} & \multicolumn{3}{c|}{21.47 / 0.6325 /   7.83} & \multicolumn{3}{c}{24.74 / 0.7260 /   4.73} \\
\multicolumn{1}{c|}{BasicVSR++ \cite{chan2022basicvsr++} + FFTformer \cite{kong2023efficient}}             & \multicolumn{3}{c|}{26.20 / 0.7746 / 0.45}   & \multicolumn{3}{c|}{22.84 / 0.6479 / 6.34}   & \multicolumn{3}{c|}{27.46 / 0.8386 / 3.65}   & \multicolumn{3}{c|}{21.34 / 0.6286 / 7.62}   & \multicolumn{3}{c}{24.46 / 0.7224 / 4.52}   \\
\multicolumn{1}{c|}{BasicVSR++ \cite{chan2022basicvsr++} + RVRT \cite{liang2022recurrent}}                  & \multicolumn{3}{c|}{26.35 / 0.7897 / 0.40}   & \multicolumn{3}{c|}{23.70 / 0.7165 / 4.36}   & \multicolumn{3}{c|}{27.74 / 0.8438 / 3.32}   & \multicolumn{3}{c|}{21.98 / 0.6905 / 5.88}   & \multicolumn{3}{c}{24.92 / 0.7604 / 3.49}   \\
\multicolumn{1}{c|}{BasicVSR++ \cite{chan2022basicvsr++} + GShiftNet \cite{li2023simple}}             & \multicolumn{3}{c|}{26.30 / 0.7862 / 0.36}   & \multicolumn{3}{c|}{23.33 / 0.6824 / 4.97}   & \multicolumn{3}{c|}{27.65 / 0.8360 / 3.38}   & \multicolumn{3}{c|}{21.69 / 0.6627 / 6.03}   & \multicolumn{3}{c}{24.74 / 0.7418 / 3.69}   \\ \hline
\multicolumn{1}{c|}{FTVSR \cite{qiu2022learning} + Restormer \cite{zamir2022restormer}}                  & \multicolumn{3}{c|}{26.31 / 0.7724 / 0.50}   & \multicolumn{3}{c|}{23.10 / 0.6533 / 6.81}   & \multicolumn{3}{c|}{27.92 / 0.8364 / 3.91}   & \multicolumn{3}{c|}{21.48 / 0.6329 / 7.83}   & \multicolumn{3}{c}{24.70 / 0.7238 / 4.76}   \\
\multicolumn{1}{c|}{FTVSR \cite{qiu2022learning} + FFTformer \cite{kong2023efficient}}                  & \multicolumn{3}{c|}{26.07 / 0.7679 / 0.51}   & \multicolumn{3}{c|}{22.83 / 0.6489 / 6.27}   & \multicolumn{3}{c|}{27.30 / 0.8328 / 3.76}   & \multicolumn{3}{c|}{21.32 / 0.6282 / 7.61}   & \multicolumn{3}{c}{24.38 / 0.7195 / 4.54}   \\
\multicolumn{1}{c|}{FTVSR \cite{qiu2022learning} + RVRT \cite{liang2022recurrent}}                       & \multicolumn{3}{c|}{26.30 / 0.7863 / 0.42}   & \multicolumn{3}{c|}{23.73 / 0.7177 / 4.37}   & \multicolumn{3}{c|}{27.61 / 0.8392 / 3.40}   & \multicolumn{3}{c|}{22.02 / 0.6923 / 5.87}   & \multicolumn{3}{c}{24.92 / 0.7589 / 3.52}   \\
\multicolumn{1}{c|}{FTVSR \cite{qiu2022learning} + GShiftNet \cite{li2023simple}}                  & \multicolumn{3}{c|}{26.26 / 0.7827 / 0.38}   & \multicolumn{3}{c|}{23.36 / 0.6845 / 4.95}   & \multicolumn{3}{c|}{27.52 / 0.8329 / 3.45}   & \multicolumn{3}{c|}{21.75 / 0.6658 / 5.99}   & \multicolumn{3}{c}{24.72 / 0.7415 / 3.69}   \\ \hline
\multicolumn{16}{c}{Single Image Deblurring + Super-Resolution}                                                                                                                                                                                                                                                     \\ \hline
\multicolumn{1}{c|}{Restormer \cite{zamir2022restormer} + Bicubic}                & \multicolumn{3}{c|}{24.04 / 0.6404 /   0.93} & \multicolumn{3}{c|}{22.96 / 0.6359 /   6.77} & \multicolumn{3}{c|}{26.47 / 0.7613 /   5.00} & \multicolumn{3}{c|}{21.44 / 0.6185 /   7.37} & \multicolumn{3}{c}{23.73 / 0.6640 /   5.02} \\
\multicolumn{1}{c|}{Restormer \cite{zamir2022restormer} + SwinIR \cite{liang2021swinir}}                 & \multicolumn{3}{c|}{24.96 / 0.7135 / 0.68}   & \multicolumn{3}{c|}{23.21 / 0.6647 / 6.14}   & \multicolumn{3}{c|}{27.43 / 0.8117 / 4.18}   & \multicolumn{3}{c|}{21.58 / 0.6442 / 6.97}   & \multicolumn{3}{c}{24.30 / 0.7085 / 4.49}   \\
\multicolumn{1}{c|}{Restormer \cite{zamir2022restormer} + HAT \cite{chen2023activating}}                    & \multicolumn{3}{c|}{25.03 / 0.7162 / 0.67}   & \multicolumn{3}{c|}{23.22 / 0.6650 / 6.12}   & \multicolumn{3}{c|}{27.50 / 0.8142 / 4.12}   & \multicolumn{3}{c|}{21.58 / 0.6444 / 6.95}   & \multicolumn{3}{c}{24.33 / 0.7100 / 4.47}   \\
\multicolumn{1}{c|}{Restormer \cite{zamir2022restormer} + BasicVSR++ \cite{chan2022basicvsr++}}             & \multicolumn{3}{c|}{25.80 / 0.7740 / 0.49}   & \multicolumn{3}{c|}{23.19 / 0.6638 / 6.12}   & \multicolumn{3}{c|}{27.78 / 0.8268 / 3.88}   & \multicolumn{3}{c|}{21.59 / 0.6472 / 6.93}   & \multicolumn{3}{c}{24.59 / 0.7280 / 4.36}   \\
\multicolumn{1}{c|}{Restormer \cite{zamir2022restormer} + FTVSR \cite{qiu2022learning}}                  & \multicolumn{3}{c|}{25.79 / 0.7709 / 0.50}   & \multicolumn{3}{c|}{23.22 / 0.6651 / 6.19}   & \multicolumn{3}{c|}{27.71 / 0.8260 / 3.91}   & \multicolumn{3}{c|}{21.61 / 0.6481 / 6.98}   & \multicolumn{3}{c}{24.58 / 0.7275 / 4.40}   \\ \hline
\multicolumn{1}{c|}{FFTformer \cite{kong2023efficient} + Bicubic}                & \multicolumn{3}{c|}{23.98 / 0.6416 / 0.87}   & \multicolumn{3}{c|}{22.82 / 0.6382 / 6.50}   & \multicolumn{3}{c|}{26.38 / 0.7610 / 4.91}   & \multicolumn{3}{c|}{21.42 / 0.6207 / 7.22}   & \multicolumn{3}{c}{23.65 / 0.6654 / 4.88}   \\
\multicolumn{1}{c|}{FFTformer \cite{kong2023efficient} + SwinIR \cite{liang2021swinir}}                 & \multicolumn{3}{c|}{24.83 / 0.7149 / 0.67}   & \multicolumn{3}{c|}{23.01 / 0.6657 / 5.93}   & \multicolumn{3}{c|}{27.28 / 0.8115 / 4.17}   & \multicolumn{3}{c|}{21.53 / 0.6462 / 6.82}   & \multicolumn{3}{c}{24.16 / 0.7096 / 4.40}   \\
\multicolumn{1}{c|}{FFTformer \cite{kong2023efficient} + HAT \cite{chen2023activating}}                    & \multicolumn{3}{c|}{24.90 / 0.7177 / 0.66}   & \multicolumn{3}{c|}{23.03 / 0.6661 / 5.90}   & \multicolumn{3}{c|}{27.37 / 0.8141 / 4.14}   & \multicolumn{3}{c|}{21.53 / 0.6464 / 6.81}   & \multicolumn{3}{c}{24.21 / 0.7111 / 4.38}   \\
\multicolumn{1}{c|}{FFTformer \cite{kong2023efficient} + BasicVSR++ \cite{chan2022basicvsr++}}             & \multicolumn{3}{c|}{25.72 / 0.7774 / 0.48}   & \multicolumn{3}{c|}{23.00 / 0.6654 / 5.92}   & \multicolumn{3}{c|}{27.61 / 0.8273 / 3.89}   & \multicolumn{3}{c|}{21.54 / 0.6492 / 6.78}   & \multicolumn{3}{c}{24.47 / 0.7298 / 4.27}   \\
\multicolumn{1}{c|}{FFTformer \cite{kong2023efficient} + FTVSR \cite{qiu2022learning}}                  & \multicolumn{3}{c|}{25.73 / 0.7742 / 0.52}   & \multicolumn{3}{c|}{23.03 / 0.6673 / 6.00}   & \multicolumn{3}{c|}{27.47 / 0.8257 / 3.95}   & \multicolumn{3}{c|}{21.56 / 0.6505 / 6.85}   & \multicolumn{3}{c}{24.45 / 0.7294 / 4.33}   \\ \hline
\multicolumn{16}{c}{Video Deblurring + Super-Resolution}                                                                                                                                                                                                                                                            \\ \hline
\multicolumn{1}{c|}{RVRT \cite{liang2022recurrent} + Bicubic}                     & \multicolumn{3}{c|}{24.03 / 0.6356 /   0.98} & \multicolumn{3}{c|}{23.33 / 0.6540 /   5.58} & \multicolumn{3}{c|}{26.51 / 0.7645 /   4.57} & \multicolumn{3}{c|}{21.91 / 0.6436 /   5.94} & \multicolumn{3}{c}{23.95 / 0.6744 /   4.27} \\
\multicolumn{1}{c|}{RVRT \cite{liang2022recurrent} + SwinIR \cite{liang2021swinir}}                      & \multicolumn{3}{c|}{25.11 / 0.7092 / 0.71}   & \multicolumn{3}{c|}{23.57 / 0.6826 / 5.03}   & \multicolumn{3}{c|}{27.33 / 0.8121 / 3.80}   & \multicolumn{3}{c|}{22.04 / 0.6688 / 5.53}   & \multicolumn{3}{c}{24.51 / 0.7182 / 3.77}   \\
\multicolumn{1}{c|}{RVRT \cite{liang2022recurrent} + HAT \cite{chen2023activating}}                         & \multicolumn{3}{c|}{25.15 / 0.7114 / 0.70}   & \multicolumn{3}{c|}{23.58 / 0.6827 / 5.01}   & \multicolumn{3}{c|}{27.40 / 0.8143 / 3.75}   & \multicolumn{3}{c|}{22.04 / 0.6688 / 5.52}   & \multicolumn{3}{c}{24.54 / 0.7193 / 3.75}   \\
\multicolumn{1}{c|}{RVRT \cite{liang2022recurrent} + BasicVSR++ \cite{chan2022basicvsr++}}                  & \multicolumn{3}{c|}{25.96 / 0.7650 / 0.58}   & \multicolumn{3}{c|}{23.56 / 0.6832 / 5.01}   & \multicolumn{3}{c|}{27.56 / 0.8237 / 3.54}   & \multicolumn{3}{c|}{22.06 / 0.6725 / 5.49}   & \multicolumn{3}{c}{24.79 / 0.7361 / 3.66}   \\
\multicolumn{1}{c|}{RVRT \cite{liang2022recurrent} + FTVSR \cite{qiu2022learning}}                       & \multicolumn{3}{c|}{25.94 / 0.7618 / 0.60}   & \multicolumn{3}{c|}{23.70 / 0.6964 / 5.34}   & \multicolumn{3}{c|}{27.49 / 0.8229 / 3.63}   & \multicolumn{3}{c|}{22.17 / 0.6848 / 6.02}   & \multicolumn{3}{c}{24.83 / 0.7415 / 3.90}   \\ \hline
\multicolumn{1}{c|}{GShiftNet \cite{li2023simple} + Bicubic}                & \multicolumn{3}{c|}{21.37 / 0.5874 / 1.28}   & \multicolumn{3}{c|}{23.20 / 0.6488 / 5.80}   & \multicolumn{3}{c|}{26.54 / 0.7650 / 4.61}   & \multicolumn{3}{c|}{21.72 / 0.6339 / 6.27}   & \multicolumn{3}{c}{23.21 / 0.6588 / 4.49}   \\
\multicolumn{1}{c|}{GShiftNet \cite{li2023simple} + SwinIR \cite{liang2021swinir}}                 & \multicolumn{3}{c|}{20.94 / 0.6163 / 1.07}   & \multicolumn{3}{c|}{23.34 / 0.6755 / 5.25}   & \multicolumn{3}{c|}{27.39 / 0.8140 / 3.85}   & \multicolumn{3}{c|}{21.80 / 0.6585 / 5.90}   & \multicolumn{3}{c}{23.37 / 0.6911 / 4.02}   \\
\multicolumn{1}{c|}{GShiftNet \cite{li2023simple} + HAT \cite{chen2023activating}}                    & \multicolumn{3}{c|}{20.99 / 0.6182 / 1.06}   & \multicolumn{3}{c|}{23.36 / 0.6757 / 5.23}   & \multicolumn{3}{c|}{27.48 / 0.8168 / 3.80}   & \multicolumn{3}{c|}{21.80 / 0.6587 / 5.89}   & \multicolumn{3}{c}{23.41 / 0.6924 / 4.00}   \\
\multicolumn{1}{c|}{GShiftNet \cite{li2023simple} + BasicVSR++ \cite{chan2022basicvsr++}}             & \multicolumn{3}{c|}{20.98 / 0.6432 / 0.88}   & \multicolumn{3}{c|}{23.35 / 0.6766 / 5.21}   & \multicolumn{3}{c|}{27.66 / 0.8278 / 3.53}   & \multicolumn{3}{c|}{21.82 / 0.6621 / 5.84}   & \multicolumn{3}{c}{23.45 / 0.7024 / 3.87}   \\
\multicolumn{1}{c|}{GShiftNet \cite{li2023simple} + FTVSR \cite{qiu2022learning}}                  & \multicolumn{3}{c|}{21.05 / 0.6439 / 0.90}   & \multicolumn{3}{c|}{23.42 / 0.6814 / 5.41}   & \multicolumn{3}{c|}{27.56 / 0.8267 / 3.57}   & \multicolumn{3}{c|}{21.87 / 0.6657 / 6.03}   & \multicolumn{3}{c}{23.47 / 0.7044 / 3.98}   \\ \hline
\multicolumn{16}{c}{Joint Video Super-Resolution and Deblurring}                                                                                                                                                                                                                                  \\ \hline
\multicolumn{1}{c|}{HOFFR \cite{fang2022high}}                              & \multicolumn{3}{c|}{- / - / -}                       & \multicolumn{3}{c|}{- / - / -}                       & \multicolumn{3}{c|}{- / - / -}                       & \multicolumn{3}{c|}{- / - / -}                       & \multicolumn{3}{c}{27.24 / 0.7870 / -}                   \\
\multicolumn{1}{c|}{Restormer$^*$ \cite{zamir2022restormer}}  & \multicolumn{3}{c|}{26.51 / 0.7551 /   0.47} & \multicolumn{3}{c|}{27.09 / 0.7695 /   3.53} & \multicolumn{3}{c|}{30.03 / 0.8579 /   2.82} & \multicolumn{3}{c|}{25.52 / 0.7573 /   4.04} & \multicolumn{3}{c}{27.29 / 0.7850 /   2.72} \\
\multicolumn{1}{c|}{GShiftNet$^*$ \cite{li2023simple}}  & \multicolumn{3}{c|}{24.66 / 0.6730 /   0.93} & \multicolumn{3}{c|}{25.66 / 0.7190 /   3.47} & \multicolumn{3}{c|}{28.05 / 0.7995 /   3.50} & \multicolumn{3}{c|}{24.69 / 0.7187 /   3.93} & \multicolumn{3}{c}{25.77 / 0.7275 /   2.96} \\
\multicolumn{1}{c|}{BasicVSR++$^*$ \cite{chan2022basicvsr++}} & \multicolumn{3}{c|}{25.90 / 0.7234 /   0.57} & \multicolumn{3}{c|}{27.07 / 0.7699 /   3.36} & \multicolumn{3}{c|}{29.67 / 0.8475 /   3.01} & \multicolumn{3}{c|}{25.58 / 0.7601 /   3.86} & \multicolumn{3}{c}{27.06 / 0.7752 /   2.70} \\
\multicolumn{1}{c|}{RVRT$^*$ \cite{liang2022recurrent}}       & \multicolumn{3}{c|}{26.84 / 0.7764 /   0.38} & \multicolumn{3}{c|}{27.76 / 0.7903 /   2.95} & \multicolumn{3}{c|}{30.66 / 0.8694 /   2.60} & \multicolumn{3}{c|}{25.93 / 0.7740 /   3.65} & \multicolumn{3}{c}{27.80 / 0.8025 /   2.40} \\ \hline
\multicolumn{1}{c|}{FMA-Net$_s$ (Ours)} & \multicolumn{3}{c|}{{\color{blue}27.08} / {\color{blue}0.7852} / {\color{blue}0.33}}               & \multicolumn{3}{c|}{{\color{blue}28.73} / {\color{blue}0.8164} / {\color{blue}2.46}}               & \multicolumn{3}{c|}{{\color{blue}30.98} / {\color{blue}0.8745} / {\color{blue}2.42}}               & \multicolumn{3}{c|}{{\color{blue}27.03} / {\color{blue}0.8089} / {\color{blue}3.10}}               & \multicolumn{3}{c}{{\color{blue}28.46} / {\color{blue}0.8212} / {\color{blue}2.08}}               \\
\multicolumn{1}{c|}{FMA-Net (Ours)}                     & \multicolumn{3}{c|}{{\color{red}\textbf{27.19}} / {\color{red}\textbf{0.7904}} / {\color{red}\textbf{0.32}}} & \multicolumn{3}{c|}{{\color{red}\textbf{29.38}} / {\color{red}\textbf{0.8308}} / {\color{red}\textbf{2.19}}}   & \multicolumn{3}{c|}{{\color{red}\textbf{31.36}} / {\color{red}\textbf{0.8814}} / {\color{red}\textbf{2.37}}} & \multicolumn{3}{c|}{{\color{red}\textbf{27.51}} / {\color{red}\textbf{0.8232}} / {\color{red}\textbf{2.79}}} & \multicolumn{3}{c}{{\color{red}\textbf{28.83}} / {\color{red}\textbf{0.8315}} /   {\color{red}\textbf{1.92}}} \\
\toprule
\end{tabular}}
\end{center}
\vspace{-5mm}
\caption{Quantitative comparison on REDS4 for $\times 4$ VSRDB. All results are calculated on the RGB channel. {\color{red}\textbf{Red}} and {\color{blue}blue} colors indicate the best and second-best performance, respectively. The superscript $^*$ indicates that the model is retrained on the REDS \cite{nah2019ntire} training dataset for VSRDB.}
\label{tab:supp_reds4_all_exp}
\end{table*}

\end{document}